\documentclass{article}
\usepackage[T1]{fontenc}
\usepackage{newunicodechar}
\newunicodechar{’}{\textquoteright}
\newunicodechar{–}{\textendash}
\usepackage{amsmath,amsfonts,bm}

\def\eqref#1{equation~\ref{#1}}

\def\1{\bm{1}}

\DeclareMathAlphabet{\mathsfit}{\encodingdefault}{\sfdefault}{m}{sl}
\SetMathAlphabet{\mathsfit}{bold}{\encodingdefault}{\sfdefault}{bx}{n}

\DeclareMathOperator*{\argmax}{arg\,max}

\usepackage[authoryear]{natbib}
\usepackage{hyperref}
\usepackage{url}
\usepackage{float}      
\RequirePackage{algorithm}
\usepackage[english]{babel}
\usepackage[authoryear]{natbib}
\usepackage{microtype}
\usepackage{graphicx}
\usepackage{caption}
\usepackage{wrapfig}
\usepackage[export]{adjustbox}
\usepackage{algorithm}
\usepackage{algpseudocode}
\usepackage{booktabs,multirow}
\usepackage[table]{xcolor}

\usepackage{listings}
\usepackage{xcolor}
\usepackage[most]{tcolorbox}
\tcbuselibrary{listings,breakable}

\title{On the Fragility of Tasks under Composition: Implicit Budgeting in Language Models}

\author{%
Chenpan Li$^{\dagger}$,
Thies Brockmoeller$^{\dagger}$, 
Stefanie Jegelka$^{\ddagger,\dagger}$,
Sharut Gupta$^{\ddagger}$\samecolorfootnote{*},
Nikolaos Karalias$^{\ddagger}$\samecolorfootnote{*}
\\
  $^\dagger$ TU Munich, $^\ddagger$MIT CSAIL \\
  \texttt{\{chenpan.li, thies.brockmoeller\}@tum.de} \quad
  \texttt{\{stefje, sharut, stalence\}@mit.edu} \\
}

\usepackage{xcolor}
\usepackage[normalem]{ulem}
\usepackage{listings}
\definecolor{codegreen}{rgb}{0,0.6,0}
\definecolor{codegray}{rgb}{0.5,0.5,0.5}
\definecolor{codepurple}{rgb}{0.58,0,0.82}
\definecolor{backcolour}{rgb}{0.95,0.95,0.92}
\definecolor{codepurple}{rgb}{0.58,0,0.82}
\definecolor{papercolor}{HTML}{0668E1}
\definecolor{darkred}{rgb}{0.68,0.05,0.0}
\definecolor{tab_blue}{RGB}{230,245,255}
\definecolor{tab_purple}{RGB}{245,230,255}
\hypersetup{
    colorlinks,
    linkcolor={papercolor!75!black},
    citecolor={papercolor!75!black},
    urlcolor={papercolor!75!black}
}

\definecolor{codeblue}{rgb}{0.25,0.5,0.5}
\definecolor{codekw}{rgb}{0.85, 0.18, 0.50}
\definecolor{mygreen}{RGB}{0,128,0}
\usepackage{ssArxiv}
\usepackage{flafter}
\usepackage{needspace}
\usepackage[section]{placeins}

\newcommand{\samecolorfootnote}[1]{\textsuperscript{\textcolor{darkred}{#1}}}

\begin{document}

\maketitle
\vspace{-0.45in}

\begin{abstract}

Modern language models are increasingly asked to handle several tasks within a single prompt. Doing so requires solving each task correctly while integrating the resulting answers into a coherent response. In this paper, we identify a fundamental limitation in this process, where models make substantially more errors when tasks are composed, even when they can solve every constituent task reliably in isolation. This degradation cannot be fully explained by the accumulation of independent errors. Across both open-weight models and the frontier model Claude Opus 5, our findings instead point to an implicit thinking budget underlying this effect. In particular, under composition, models use substantially fewer tokens per task than in isolation, even when the full sequence remains well within the context window. Tasks that require longer solution traces in isolation are consequently more \emph{fragile} under composition, i.e., models make more errors on prompts containing these tasks. In isolation, non-fragile tasks receive more reasoning than they need and waste compute. However, by grouping tasks according to their fragility, the implicit budgeting can be manipulated to reduce token usage by $43$\% across a range of math and linguistic tasks while also preserving accuracy.\looseness=-1

Project page: \url{https://chenpanli.github.io/fragility} \\ Code: \url{https://github.com/chenpanli/fragility}
\end{abstract}
\footnotetext{denotes equal advisory role.}

\begin{figure}[!htb]
    \centering
    \includegraphics[width=0.80\linewidth]{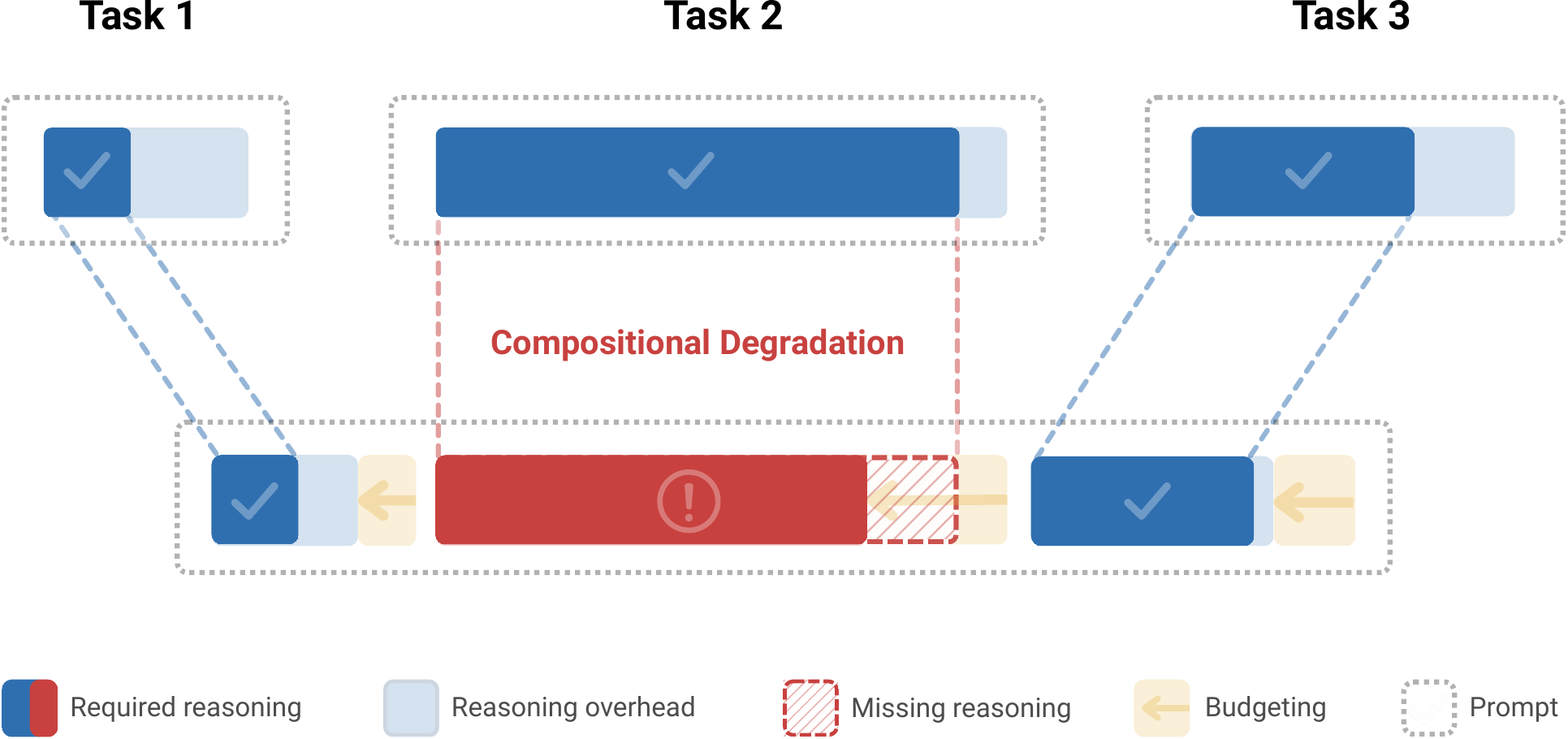}
    \caption{\textbf{Composition reduces every task's reasoning by the same factor.} When solved in isolation (top), each task receives more reasoning than it requires. When composed into a single prompt (bottom), the reasoning allocated to each task shrinks proportionally (yellow). Tasks with sufficient slack absorb this reduction and are still solved correctly, whereas the task that requires long reasoning falls below its requirement and thus fails.}
    \label{fig:mainfigure}
\end{figure}
\clearpage

\section{Introduction}
\label{sec:introduction}

A single prompt often asks a language model to do several things at once, such as answering a handful of questions, solving a set of math problems, or following a list of instructions before writing one response. Common benchmarks, however, still test models one task at a time~\citep{rein2023gpqagraduatelevelgoogleproofqa, hendrycksmath2021, gema2025mmlu}.
As scores on these benchmarks saturate, they may overstate how robust the underlying capabilities are. We ask \emph{whether a model that reliably solves a task on its own can still solve it when other tasks share the prompt}.\looseness=-1

One natural explanation is error accumulation. If each task fails with nonzero probability, solving many at once fails more often. Prior work has found this phenomenon when models are tasked to do familiar reasoning steps~\citep{press2023compositionalitygap, dziri2023faithfatelimitstransformers, sinha2026illusiondiminishingreturnsmeasuring} or follow several instructions at once~\citep{harada2025instructions, vasileva2026large}. However, accumulation is only half of the story. Because composition in these settings also changes the structure and token demands of the task, it remains unclear whether the model's ability to solve each constituent task degrades as well, and what mechanism drives that degradation.\looseness=-1

To investigate the question, we analyze the effect of composition on \emph{each constituent task} of a composition. We start by establishing that this per-task degradation exists at all. For that, we select tasks the model solves correctly in isolation, which establishes that the requisite capability is present. We then leave those tasks unchanged and simply concatenate them within a single prompt. This design separates degradation under increased task load from baseline solvability on the constituent tasks. \looseness=-1

Across state-of-the-art open models, as well as the frontier model Claude Opus~5, we find substantial performance degradation when more previously solved tasks are concatenated within the same prompt. The drop in accuracy is substantially larger than what the constituent tasks’ isolated error rates would predict, and is consistent across both mathematical and linguistic domains. We characterize this degradation along two dimensions. 1) Failures concentrate on a subset of \emph{fragile tasks} that, for a given model, remains stable across sampling noise and composition sizes. These tasks also have longer reasoning traces in isolation. 2) As more tasks are presented together, models systematically reduce the reasoning tokens devoted to each task, even though no explicit per-task token constraint is imposed. This reduction follows a clear scaling pattern, with reasoning per task decreasing approximately as a \emph{power law} with composition size.
Together, these findings reveal an implicit thinking budget under composition, under which fragile tasks disproportionately lose accuracy as per-task reasoning decreases.\looseness=-1

This mechanism also creates an opportunity for more efficient inference than solving every task separately. Non-fragile tasks can be grouped into larger batches, where the implicit thinking budget reduces per-task reasoning and token usage without impacting accuracy. Fragile tasks, in contrast, are placed in smaller batches so that each receives more reasoning and remains accurate. We demonstrate this with fragility-aware batching, a strategy that composes each task based on its fragility. By using empirical measurements of the fragility of tasks, it reduces token usage by $43$\% on average with no drop in performance compared to solving each task individually. Because it is entirely black-box, it also applies directly to proprietary models, reducing generated tokens by $24\%$ on GPQA-Diamond with Claude Opus~5 without degrading performance.\looseness=-1

Taken together, these results show that the failure of LLMs to generalize their reasoning budget management is a double-edged sword. Under composition, it causes fragile tasks to fail, whereas in isolation, it makes models overspend reasoning on non-fragile tasks, thus wasting compute during inference. Our key contributions and findings are:\looseness=-1

\begin{itemize}
    \item We show that simply concatenating tasks that are already solved correctly in isolation causes substantial accuracy degradation across both frontier and open models (Section~\ref{sec:results}).\looseness=-1
    \item We identify an implicit per-task thinking budget that shrinks as a power law in the number of composed tasks (Section~\ref{sec:budget}).\looseness=-1
    \item We show that this uniform reduction concentrates degradation on tasks with long isolated reasoning, a set that is stable within each model but differs between models (Sections~\ref{sec:budget} and~\ref{sec:transfer-fragility-set}). We refer to this set of tasks as \emph{fragile}.\looseness=-1
    \item We turn these findings into fragility-aware batching, a strategy that groups tasks according to their measured fragility. It reduces generated tokens by $\sim$43\% across all datasets while preserving accuracy on both state-of-the-art open-weight models and proprietary frontier models (Section~\ref{sec:repair}).\looseness=-1
\end{itemize}

\section{Preliminaries}
\label{sec:methods}

Our central goal is to investigate whether an LLM that can solve a task in isolation keeps this ability when several such tasks are presented together. To answer this, we use a controlled setting in which the constituent tasks are unchanged and only the number of tasks in the prompt varies. We then measure how \emph{each task’s accuracy changes} relative to its performance in isolation. All experiments share the setup described below.\looseness=-1

\Needspace{4\baselineskip}
\noindent\textbf{Compositions.} We compose independent tasks drawn from the same dataset $\mathcal{D}$. A composition of size $k$ is a tuple $C = (t_1, \dots, t_k)$ with $t_i \in \mathcal{D}$, rendered into a single prompt by a fixed per-dataset template that asks the model to answer all $k$ tasks in order.
We split the model's single response into $k$ answers and grade each for correctness (further details in Appendix~\ref{app:datasets}).\looseness=-1

\Needspace{4\baselineskip}
\noindent\textbf{Accuracy drop.} For a model $m$, the accuracy drop of each individual task $t_i$ at composition size $k$ is $\Delta^{m}_i(k) = \mathrm{acc}^{m}_i(k) - \mathrm{acc}^{m}_i(1)$, where negative values indicate degradation. Unless stated otherwise, accuracy is estimated from 5 independent trials.\looseness=-1

\Needspace{4\baselineskip}
\noindent\textbf{Datasets.} We use eight datasets whose tasks vary widely in the reasoning they require. We build two datasets ourselves: 1,000 indefinite integrals with randomized coefficients across diverse function classes, verified symbolically, and OneWordQA, a set of factual questions with single-word or numeric answers that require minimal reasoning. The remaining six are established benchmarks: GPQA-Diamond~\citep{rein2023gpqagraduatelevelgoogleproofqa}, GSM8K~\citep{cobbe2021trainingverifierssolvemath}, MATH-500~\citep{hendrycksmath2021, lightman2023lets}, MMLU-Redux~\citep{gema2025mmlu}, StrategyQA~\citep{geva2021didaristotleuselaptop} and AIME (further details in Appendix~\ref{app:datasets}).\looseness=-1

\Needspace{4\baselineskip}
\noindent\textbf{Models.} We evaluate three Qwen3 models (4B Thinking, 8B, 14B)~\citep{yang2025qwen3technicalreport}, trained by strong-to-weak distillation from larger RL-trained Qwen3 models, and Qwen3.8 27B~\citep{qwen38}. We also evaluate three DeepSeek-R1-distilled models (Qwen 7B, Llama 8B, Qwen3 8B)~\citep{deepseekai2025deepseekr1incentivizingreasoningcapability}, obtained by supervised fine-tuning on R1 reasoning traces, and the frontier model Claude Opus~5 with medium reasoning effort.
The first two groups let us compare two different post-training recipes at similar scale. Open-weight models are sampled at temperature 0.6 and use a context limit of at least $131{,}072$ tokens (further details in Appendix~\ref{app:sampling-llms}).\looseness=-1

\section{Composing tasks causes performance degradation}
\label{sec:results}

Prior work has repeatedly reported that accuracy falls when several tasks share a single prompt.
However, these evaluations do not isolate whether a model fails because it cannot solve a task reliably or purely because the task is composed with others, which leaves the underlying mechanism unclear.\looseness=-1

Building on these observations, we ask whether degradation can arise from composition alone. Answering this requires separating a model's ability to solve a task from its ability to solve the same task in composition. For that, we restrict each model $M$ to a subset $\mathcal{D}_M \subseteq \mathcal{D}$ of \emph{confidently solved} tasks: a task enters $\mathcal{D}_M$ only if $M$ solves it in isolation in three out of three independent attempts. We then build compositions $C$ of size $k \in \{3, 6, 10, 20\}$\footnote{$k=20$ is left out on GPQA-Diamond and AIME because of overflowing context.} exclusively from $\mathcal{D}_M$, sampling task categories uniformly within each dataset, and evaluate the accuracy of each task within them. The accuracies at $k=1$ come from runs independent of those used for selection.\looseness=-1

\begin{figure}[!htbp]
  \centering
  \includegraphics[width=0.49\linewidth, valign=t]{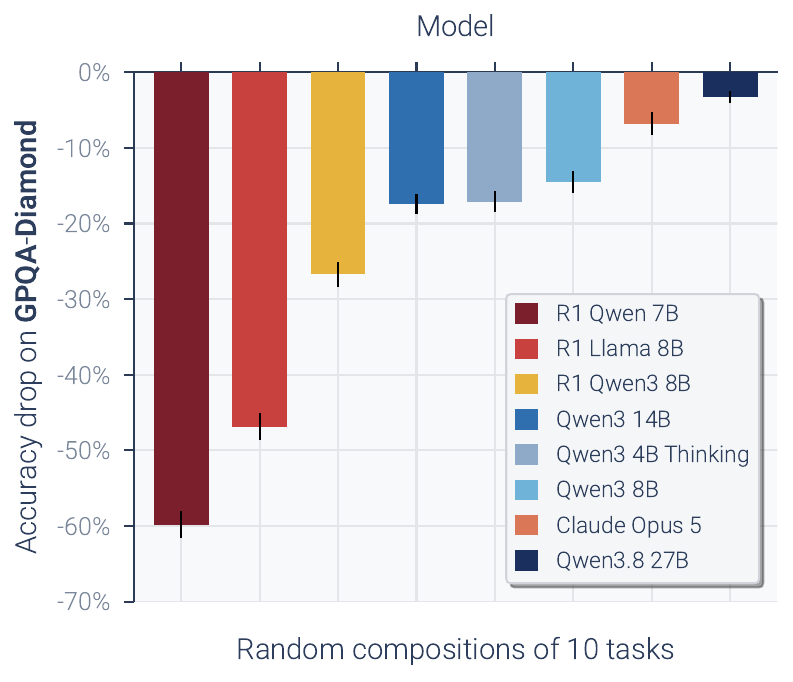}\hfill
  \includegraphics[width=0.49\linewidth, valign=t]{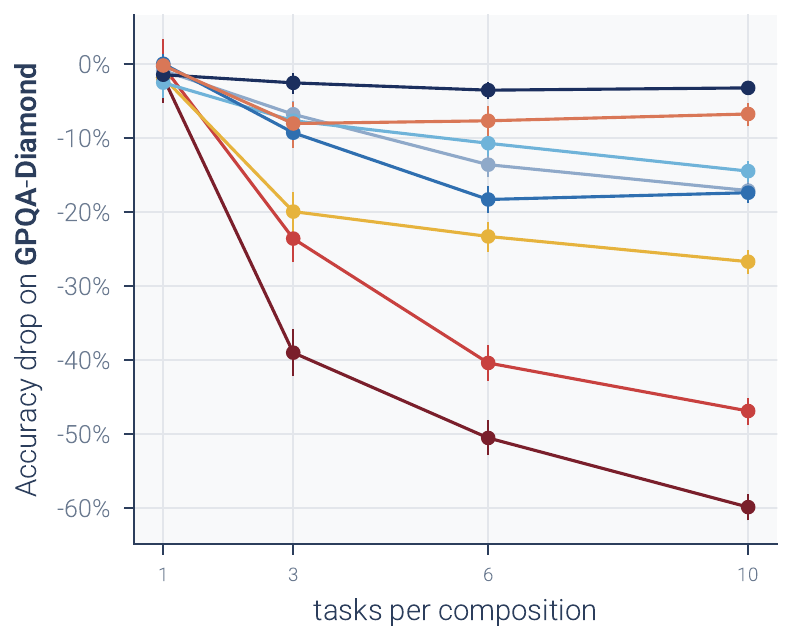}
  \caption{Accuracy degradation on confidently solved GPQA-Diamond tasks. \textbf{Left:} accuracy drop per model at $k=10$. \textbf{Right:} accuracy drop as a function of composition size $k$. All tasks were solved in three out of three isolated attempts, yet accuracy drops once they are composed.}
  \label{fig:success-conditioned-analysis}
\end{figure}

\FloatBarrier
\Needspace{18\baselineskip}
\begin{wrapfigure}[17]{r}{0.36\textwidth}
  \centering
  \includegraphics[width=\linewidth]{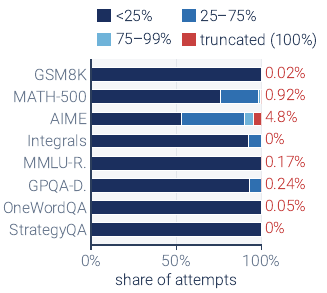}
  \caption{Context usage across datasets. Most responses use under 25\% of the context window; in most datasets, under 1\% hit the limit.\looseness=-1}
  \label{fig:context-usage}
\end{wrapfigure}

As Figure~\ref{fig:success-conditioned-analysis} shows, simply concatenating tasks that are solvable in isolation causes substantial accuracy degradation across all models on GPQA-Diamond~\citep{rein2023gpqagraduatelevelgoogleproofqa}. This observation also holds for all other datasets we tested, except OneWordQA and StrategyQA, whose tasks require little reasoning. Furthermore, this degradation cannot be explained by context-length issues, since nearly all responses stay within the context limit (see Figure~\ref{fig:context-usage} and Appendix~\ref{app:success-conditioned-analysis}). Compositional degradation is therefore driven by effects inherent to composition and cannot be predicted from the solvability of tasks in isolation alone.\looseness=-1

The extent of this degradation varies strongly across models. The R1-distilled models degrade the most, while the Qwen3 models degrade moderately less. Frontier models are not immune either: Claude Opus 5, despite its leading benchmark results, loses about $7$\,pp at $k=10$ on GPQA-Diamond, on tasks it solved in all three runs individually.\looseness=-1

Contrary to explanations for compositions with dependent tasks, where models either match solution patterns seen in training instead of learning systematic procedures~\citep{dziri2023faithfatelimitstransformers} or fail to combine facts they know individually~\citep{press2023compositionalitygap}, we propose that compositional degradation arises from a misallocation of the implicit thinking budget.\looseness=-1

\subsection{Degradation as a misallocation of the implicit thinking budget}
\label{sec:budget}

Reasoning models solve tasks by thinking stepwise, a process shown to be crucial for task success~\citep{wei2023chainofthoughtpromptingelicitsreasoning}. We therefore ask whether the accuracy degradation under composition can be explained by changes in thinking behavior. A central property of this behavior is the number of thinking tokens the model spends on a task, i.e., how much effort it invests in it. This effort can strongly affect whether a task is solved~\citep{ snell2024scalingllmtesttimecompute}, and prior work has shown that overall reasoning shortens when batching tasks~\citep{srivastava2026constraint, deng2026conpress, cheng2023batchpromptingefficientinference}. What remains unclear is how each constituent task is affected, that is, how its thinking length under composition compares to its thinking length in isolation and how that change affects the task's success.\looseness=-1

\begin{figure}[!htbp]
  \centering
  \includegraphics[width=\linewidth, valign=t]{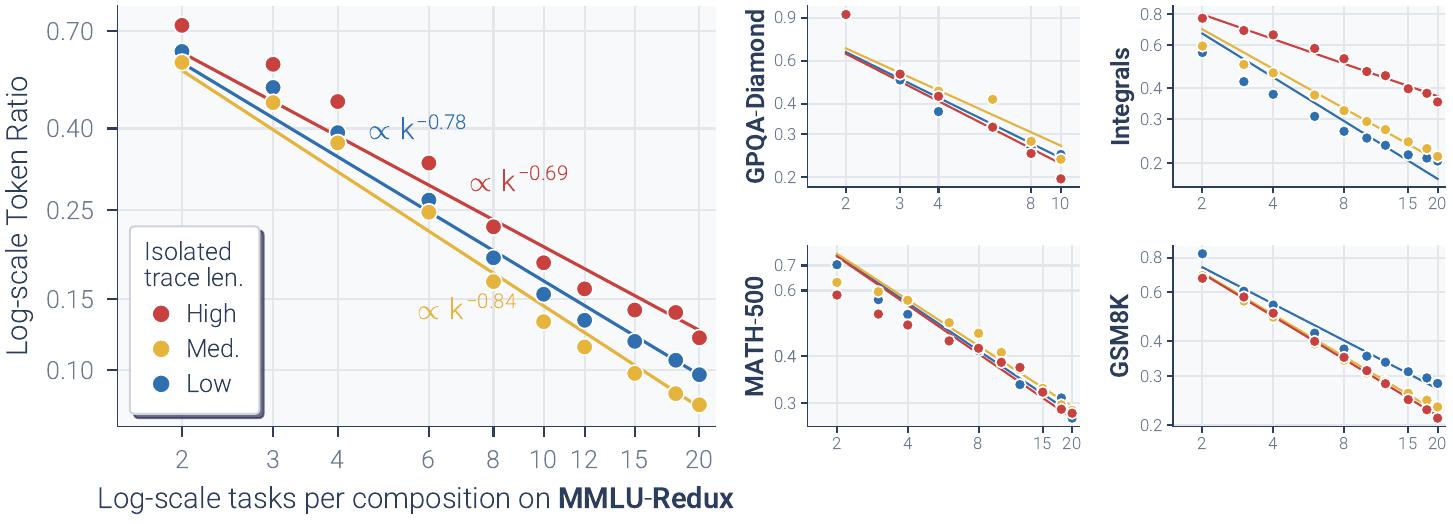}
  \caption{Token ratio $r_i(k) = T_i(k)/T_i(1)$ as a function of composition size $k$, where $T_i(k)$ is the number of thinking tokens spent on task $t_i$ within a composition of size $k$ and $T_i(1)$ the number spent in isolation. Per-task token counts in compositions are obtained by locating task tags in the reasoning trace. Tasks are binned by the quantile of their isolated trace length into Low ($\leq 20\%$), Medium ($20$--$80\%$), and High ($> 80\%$). On log-log axes, the ratio decreases approximately linearly in $\log k$, indicating that LLMs shrink the relative token usage of each task under composition by roughly $k^{m}$, with no systematic dependence on $T_i(1)$. Results are from Qwen3 4B Thinking.}
  \label{fig:k-vs-token-ratio}
\end{figure}
\Needspace{4\baselineskip}
\noindent\textbf{Thinking traces shrink under composition.}
We close this gap by measuring the thinking tokens spent on each task within a composition of $k$ tasks, denoted $T_i(k)$. To this end, we prefix each task with a unique tag (Q01, Q02, \dots) in the prompt that does not appear in the task description itself. These tags let us attribute each segment of the reasoning trace to the task it addresses (details in Appendix~\ref{app:trace-segmentation}).
We quantify the reduction for each task by the token ratio $r_i(k) = T_i(k)/T_i(1)$ between its thinking tokens under composition and in isolation. Figure~\ref{fig:k-vs-token-ratio} shows how this token ratio evolves with composition size. With one exception (R1 Qwen 7B on MATH-500), the thinking tokens of almost all tasks decrease monotonically as compositions grow (further models and datasets in Appendix~\ref{app:composition-size-token-ratio}); a weak positional bias is discussed in Appendix~\ref{app:positional-bias}.\looseness=-1

\begin{figure}[!t]
  \centering
  \includegraphics[width=0.49\linewidth, valign=t]{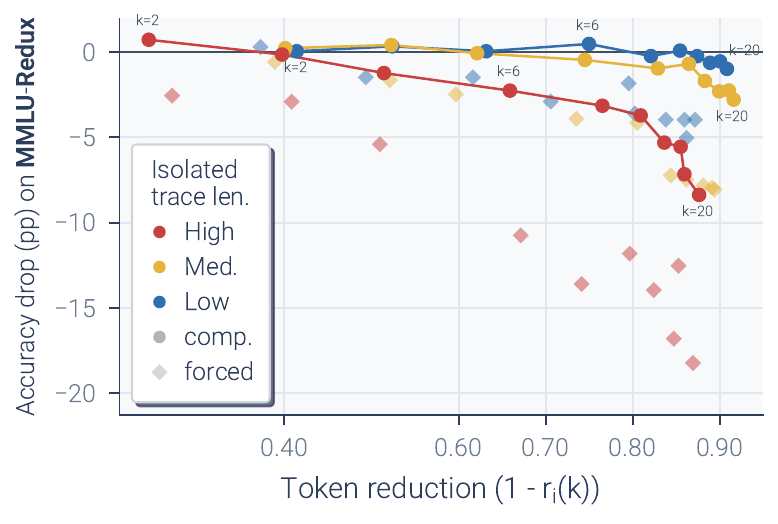}\hfill
  \includegraphics[width=0.49\linewidth, valign=t]{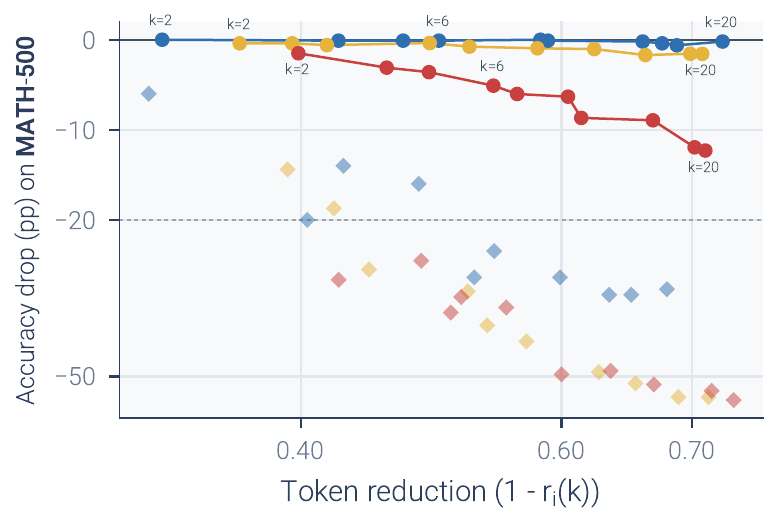}
  \caption{Accuracy drop as a function of the relative thinking token reduction for Qwen3 4B Thinking on MMLU-Redux (left) and MATH-500 (right) below the dashed line, the y-axis is logarithmic for presentation purposes. Tasks are binned by the quantile of their isolated trace length into Low ($\leq 20\%$), Medium ($20$--$80\%$), and High ($> 80\%$). As the reduction through composition (comp.) grows, accuracy drops, and tasks with high reasoning requirements degrade substantially more than low-effort tasks. Forcing a fixed budget externally (forced) generally causes stronger degradation, but the degradation follows the same patterns (more details in Appendix~\ref{app:token-ratio-accuracy}).}
  \label{fig:token-ratio-vs-accuracy}
\end{figure}

\Needspace{4\baselineskip}
\noindent\textbf{Long-reasoning tasks degrade more.}
This reduction has important implications, since prior work has established that the number of thinking tokens spent on a task can influence its accuracy. In particular, interrupting the thinking process externally at earlier stages causes a larger degradation~\citep{muennighoff2025s1simpletesttimescaling}. Composing tasks has an analogous effect on thinking length, as it implicitly constrains the per-task thinking tokens but without any external enforcement. This naturally raises the question of whether the implicit token reduction under composition is related to the accuracy degradation.
Figure~\ref{fig:token-ratio-vs-accuracy} shows how the reduction affects tasks with different isolated thinking lengths. Tasks with longer isolated traces degrade substantially more when thinking tokens are reduced, whereas tasks with shorter traces are largely unaffected. We refer to tasks that break under composition in this way as \emph{fragile}.\looseness=-1

\Needspace{4\baselineskip}
\noindent\textbf{Thinking length follows a power law.}
One possible source of the uneven degradation in Figure~\ref{fig:token-ratio-vs-accuracy} could be that long-reasoning tasks lose a larger share of their thinking tokens.
The token ratio approximately follows a \emph{power law in $k$},
\begin{equation}
  \log r_i(k) \approx m \log k
  \quad\Longleftrightarrow\quad r_i(k) = \frac{T_i(k)}{T_i(1)} \approx k^{m}.
  \label{eq:power-law}
\end{equation}
Since $r_i(1) = 1$ by definition, we fit a line through the origin on log-log axes for each model and dataset. The fitted slopes have a median of $m=-0.54$ ($m \in [-0.88, -0.04]$), with a median $R^2$ of $0.91$; averaged per model, $R^2$ ranges from $0.55$ to $0.94$ (Appendix~\ref{app:composition-size-token-ratio}). Under composition, the reasoning trace of each task thus shrinks by approximately the same factor $k^{m}$, without systematic dependence on its isolated length $T_i(1)$, so long-reasoning tasks do not systematically lose a larger share of their reasoning than short ones. The model preserves the relative allocation across tasks, assigning more reasoning to tasks that required more in isolation, but applies roughly the same proportional reduction to all of them.
Considering that every task loses roughly the same fraction of its reasoning, the uneven degradation is best explained by how much reasoning each task can afford to lose. Short isolated traces seem to contain redundancy in reasoning beyond what the task requires, whereas long traces lie closer to the minimum reasoning needed to solve the task reliably. Hence, the same relative reduction pushes long-reasoning tasks below their minimal requirement while leaving short ones above it, concentrating degradation on tasks with long isolated traces. Since $k^{m}$ decreases with $k$ ($m<0$), the implicit thinking budget shrinks as more tasks are composed, and degradation grows accordingly. This reveals a systematic failure of implicit thinking budget allocation under composition. The way models allocate their implicit thinking budget \emph{does not generalize well to composed tasks}.\looseness=-1

\Needspace{4\baselineskip}
\subsection{Fragile tasks are stable within a model}
\label{sec:transfer-fragility-set}
It remains to establish whether the implicit thinking budget estimation fails consistently on the same tasks across models and composition sizes. This tells us how the proposed misallocation changes as the conditions, i.e., model and composition size, vary.
Since a failure in implicit thinking budget estimation manifests as accuracy degradation, we can test this consistency directly on the accuracy drops. For two conditions $A$ and $B$, we obtain the accuracy drops across tasks, $\Delta_A$ and $\Delta_B$, and measure their Pearson correlation, applying Spearman's correction for attenuation to reduce the impact of noisy accuracy estimates (details in Appendix~\ref{app:degradation-correlation}).\looseness=-1

At a fixed composition size, the mean correlation across models is only $0.24\pm0.03$. Different models thus largely disagree on which tasks degrade, even when evaluated under identical conditions. The exception is models trained with the same recipe, such as Qwen3 8B and 14B, which correlate at $0.56\pm0.09$. Within a single model, the correlation across composition sizes reaches $0.81\pm0.04$. When the same model faces compositions of different sizes, groupings and orderings, the same set of fragile tasks tends to break.
Fragility is therefore a stable property of a task for a given model. Measuring a task's accuracy drop at some composition sizes lets us anticipate how it will degrade in new compositions presented to the same model, which makes fragility predictable.\looseness=-1

\section{Exploiting implicit budgeting for efficient inference}
\label{sec:repair}

\providecommand{\relcell}[2]{{\small #1}\,{\scriptsize$(#2\%)$}}
\providecommand{\ppcell}[2]{{\small #1}\,{\scriptsize$(#2)$}}
\begin{table}[!htbp]
\centering
\caption{Task accuracy (Acc.) and output tokens (Tok., includes thinking and answer) per task under different batching policies. For each policy, the left column reports accuracy and the right column tokens, with the change to solving in isolation ($k=1$) in parentheses, given as an absolute difference in percentage points for accuracy and as a relative change in \% for tokens. The Average row gives the mean change over the models shown. Standard fixed-size random batching causes substantial accuracy degradation, which fragility-aware batching largely mitigates. With a model-specific linear predictor, part of these token savings carries over to unseen tasks while still avoiding compositional degradation.}
\label{tab:batching_results}
\resizebox{\textwidth}{!}{%
\begin{tabular}{lllll|llll}
\toprule
 & \multicolumn{2}{c}{Isolated, $k{=}1$} & \multicolumn{2}{c|}{Random batching, $k{=}6$} & \multicolumn{2}{c}{Fragility-aware (measured)} & \multicolumn{2}{c}{Fragility-aware (predictor)} \\
\cmidrule(lr){2-3} \cmidrule(lr){4-5} \cmidrule(lr){6-7} \cmidrule(lr){8-9}
 & \multicolumn{1}{c}{Acc.} & \multicolumn{1}{c}{Tok.} & \multicolumn{1}{c}{Acc.} & \multicolumn{1}{c|}{Tok.} & \multicolumn{1}{c}{Acc.} & \multicolumn{1}{c}{Tok.} & \multicolumn{1}{c}{Acc.} & \multicolumn{1}{c}{Tok.} \\
\midrule
\multicolumn{9}{l}{\textit{GPQA-Diamond}} \\
\midrule
Qwen3 4B T. & 66.7\% & 7663 & \ppcell{56.4\%}{-10.3} & \relcell{2470}{-67.8} & \ppcell{66.8\%}{+0.2} & \relcell{5200}{-32.1} & \ppcell{67.3\%}{+0.7} & \relcell{7534}{-1.7} \\
Qwen3 8B & 58.7\% & 5973 & \ppcell{53.4\%}{-5.3} & \relcell{1950}{-67.4} & \ppcell{60.0\%}{+1.3} & \relcell{3520}{-41.1} & \ppcell{56.9\%}{-1.8} & \relcell{5106}{-14.5} \\
Qwen3 14B & 61.5\% & 4597 & \ppcell{51.8\%}{-9.7} & \relcell{1500}{-67.4} & \ppcell{57.9\%}{-3.7} & \relcell{2111}{-54.1} & \ppcell{61.3\%}{-0.2} & \relcell{3665}{-20.3} \\
Qwen3.8 27B & 83.2\% & 4303 & \ppcell{79.2\%}{-4.0} & \relcell{2740}{-36.3} & \ppcell{84.0\%}{+0.8} & \relcell{4102}{-4.7} & \ppcell{85.3\%}{+2.1} & \relcell{4324}{+0.5} \\
R1 Qwen 7B & 49.1\% & 6686 & \ppcell{23.6\%}{-25.5} & \relcell{1305}{-80.5} & \ppcell{49.6\%}{+0.5} & \relcell{4037}{-39.6} & \ppcell{46.7\%}{-2.4} & \relcell{5953}{-11.0} \\
R1 Llama 8B & 47.7\% & 7424 & \ppcell{32.1\%}{-15.6} & \relcell{1771}{-76.2} & \ppcell{49.6\%}{+1.9} & \relcell{4526}{-39.0} & \ppcell{48.6\%}{+0.9} & \relcell{5816}{-21.7} \\
R1 Qwen3 8B & 60.8\% & 9084 & \ppcell{48.1\%}{-12.6} & \relcell{2398}{-73.6} & \ppcell{61.9\%}{+1.2} & \relcell{6595}{-27.4} & \ppcell{59.8\%}{-1.0} & \relcell{8877}{-2.3} \\
Claude Opus 5 & 89.5\% & 689 & \ppcell{88.7\%}{-0.9} & \relcell{564}{-18.2} & \ppcell{89.2\%}{-0.4} & \relcell{521}{-24.3} & \ppcell{87.6\%}{-2.0} & \relcell{686}{-0.5} \\
\midrule[0.08em]
\rowcolor{gray!12}
Average & 64.7\% & 5802 & $-10.5$ & \multicolumn{1}{l}{$-60.9\%$} & \multicolumn{1}{|l}{$+0.2$} & $-32.8\%$ & $-0.5$ & $-8.9\%$ \\
\midrule
\multicolumn{9}{l}{\textit{MATH-500}} \\
\midrule
Qwen3 4B T. & 97.6\% & 4064 & \ppcell{96.4\%}{-1.1} & \relcell{2148}{-47.2} & \ppcell{97.7\%}{+0.1} & \relcell{3716}{-8.6} & \ppcell{96.8\%}{-0.8} & \relcell{3605}{-11.3} \\
Qwen3 8B & 95.9\% & 3449 & \ppcell{95.6\%}{-0.3} & \relcell{1751}{-49.2} & \ppcell{95.1\%}{-0.8} & \relcell{2173}{-37.0} & \ppcell{95.9\%}{+0.1} & \relcell{2775}{-19.5} \\
Qwen3 14B & 96.9\% & 2604 & \ppcell{95.9\%}{-1.0} & \relcell{1375}{-47.2} & \ppcell{96.6\%}{-0.2} & \relcell{2454}{-5.8} & \ppcell{96.6\%}{-0.2} & \relcell{2444}{-6.2} \\
Qwen3.8 27B & 98.1\% & 1014 & \ppcell{96.7\%}{-1.5} & \relcell{748}{-26.2} & \ppcell{97.8\%}{-0.4} & \relcell{943}{-7.1} & \ppcell{97.2\%}{-0.9} & \relcell{994}{-2.0} \\
R1 Qwen 7B & 93.0\% & 2383 & \ppcell{69.8\%}{-23.2} & \relcell{1658}{-30.4} & \ppcell{93.2\%}{+0.2} & \relcell{2205}{-7.5} & \ppcell{93.4\%}{+0.4} & \relcell{2381}{-0.1} \\
R1 Llama 8B & 88.7\% & 3019 & \ppcell{79.4\%}{-9.3} & \relcell{1569}{-48.0} & \ppcell{88.5\%}{-0.2} & \relcell{2948}{-2.3} & \ppcell{88.8\%}{+0.2} & \relcell{3333}{+10.4} \\
R1 Qwen3 8B & 96.8\% & 6767 & \ppcell{93.0\%}{-3.8} & \relcell{3705}{-45.2} & \ppcell{96.6\%}{-0.2} & \relcell{6608}{-2.4} & \ppcell{96.7\%}{-0.2} & \relcell{6657}{-1.6} \\
\midrule[0.08em]
\rowcolor{gray!12}
Average & 95.3\% & 3329 & $-5.7$ & \multicolumn{1}{l}{$-41.9\%$} & \multicolumn{1}{|l}{$-0.2$} & $-10.1\%$ & $-0.2$ & $-4.3\%$ \\

\midrule
\multicolumn{9}{l}{\textit{MMLU-Redux}} \\
\midrule
Qwen3 4B T. & 84.7\% & 1944 & \ppcell{82.9\%}{-1.8} & \relcell{527}{-72.9} & \ppcell{85.3\%}{+0.6} & \relcell{527}{-72.9} & \ppcell{84.9\%}{+0.2} & \relcell{1663}{-14.5} \\
Qwen3 8B & 87.2\% & 921 & \ppcell{84.0\%}{-3.3} & \relcell{306}{-66.7} & \ppcell{86.1\%}{-1.1} & \relcell{285}{-69.0} & \ppcell{86.9\%}{-0.3} & \relcell{922}{+0.2} \\
Qwen3 14B & 89.0\% & 718 & \ppcell{85.8\%}{-3.3} & \relcell{212}{-70.5} & \ppcell{88.2\%}{-0.9} & \relcell{319}{-55.6} & \ppcell{88.3\%}{-0.7} & \relcell{678}{-5.5} \\
Qwen3.8 27B & 92.4\% & 488 & \ppcell{91.9\%}{-0.5} & \relcell{262}{-46.4} & \ppcell{91.8\%}{-0.6} & \relcell{196}{-59.8} & \ppcell{92.1\%}{-0.3} & \relcell{352}{-27.9} \\
R1 Qwen 7B & 70.1\% & 1079 & \ppcell{60.6\%}{-9.5} & \relcell{397}{-63.2} & \ppcell{69.7\%}{-0.3} & \relcell{601}{-44.3} & \ppcell{70.4\%}{+0.3} & \relcell{1127}{+4.5} \\
R1 Llama 8B & 75.5\% & 1458 & \ppcell{58.6\%}{-16.9} & \relcell{431}{-70.4} & \ppcell{75.9\%}{+0.3} & \relcell{896}{-38.5} & \ppcell{75.3\%}{-0.2} & \relcell{1430}{-1.9} \\
R1 Qwen3 8B & 81.6\% & 1711 & \ppcell{64.5\%}{-17.1} & \relcell{279}{-83.7} & \ppcell{82.1\%}{+0.5} & \relcell{1273}{-25.6} & \ppcell{82.2\%}{+0.6} & \relcell{1714}{+0.2} \\
\midrule[0.08em]
\rowcolor{gray!12}
Average & 82.9\% & 1188 & $-7.5$ & \multicolumn{1}{l}{$-67.7\%$} & \multicolumn{1}{|l}{$-0.2$} & $-52.2\%$ & $-0.1$ & $-6.4\%$ \\
\bottomrule
\end{tabular}}
\end{table}

So far, our results suggest that LLMs fail on compositions mainly because their estimate of the implicit thinking budget does not generalize, rather than because they cannot transfer their problem-solving skills. If so, controlling how much reasoning each task receives should prevent the drop. The composition size $k$ offers a practical control mechanism on implicit budgeting, since the token reduction ratio follows a power law in $k$. Hence, assigning robust tasks to larger compositions and fragile tasks to smaller ones could save tokens while limiting accuracy degradation.\looseness=-1

\subsection{Fragility-aware batching}
\label{sec:oracle-fix}
To test whether this idea is viable, we first ask how many tokens could be saved if each task's fragility were known. We estimate fragility from the measured accuracy drops and compose tasks accordingly, which gives a sense of what fragility-aware batching can achieve. In Section~\ref{sec:predict-fragility}, we replace these measurements with a linear predictor that predicts fragility, turning the approach into a method that can be deployed.
Formally, we seek to assign each task a composition size $k_i$ that maximizes the expected token savings while bounding the expected accuracy degradation by $B$. We express this trade-off as a Lagrangian relaxation,
\begin{equation}
k_i(\lambda) = \argmax_{k \in \mathcal{K}}
\Big[ \mathbb{E}[\Delta_i(k)] + \lambda \big(1 - \mathrm{cost}_k\big) \Big],
\label{eq:lagrangian}
\end{equation}
where $\mathcal{K}$ is the set of available composition sizes, $\mathbb{E}[\Delta_i(k)]$ is the expected accuracy drop of task $t_i$ at composition size $k$, and $\mathrm{cost}_k \in (0, 1]$ is the model-specific token cost of solving a task within a composition of size $k$, relative to solving it in isolation. The parameter $\lambda \ge 0$ acts as an exchange rate between saved tokens and accuracy.
We then choose $\lambda$ such that the mean expected accuracy drop over the dataset, $\bar{\Delta}(\lambda) = \frac{1}{n}\sum_{i=1}^{n} \mathbb{E}\big[\Delta_i(k_i(\lambda))\big]$, conservatively stays within the budget $B$,
\begin{equation}
\lambda^{*} = \max \Big\{ \lambda \ge 0 \;:\;
\bar{\Delta}(\lambda) - \mathrm{SE}\big(\bar{\Delta}(\lambda)\big) \ge -B \Big\}.
\label{eq:lambda}
\end{equation}
We find $\lambda^{*}$ by a grid search on held-out splits, using repeated nested $5$-fold cross-validation \citep{cawley2010onoverfitting} to reduce the effect of noise in the empirical accuracy estimates. Given the assignments $k_i(\lambda^{*})$, we greedily group tasks into as few compositions as possible such that every task $t_i$ lands in a composition $C$ with $|C| \le k_i(\lambda^{*})$ (details in Appendix~\ref{app:fragility-aware-batching}). Throughout our experiments, we use the composition sizes $\mathcal{K} = \{1, 3, 6, 10, 20\}$, where $k=1$ solves a task alone, and an accuracy drop budget of $B = 0.5\%$.\looseness=-1

Table~\ref{tab:batching_results} compares the accuracy degradation and token savings of different batching strategies relative to solving tasks in isolation. The common practice of batching tasks randomly into fixed-size compositions~\citep{cheng2023batchpromptingefficientinference}, here of six tasks, saves the most tokens on average but degrades accuracy strongly, particularly on GPQA-Diamond, whose tasks require many reasoning steps. Fragility-aware batching almost entirely avoids this degradation while still achieving substantial token savings. Averaged over all models and over every dataset but AIME, where fragility-aware batching leaves nearly every task in isolation (Appendix~\ref{app:batching-full}), it saves $43\%$ of tokens while keeping accuracy constant ($+0.6$\,pp). How these savings arise depends on the dataset. On OneWordQA and StrategyQA, fragility-aware batching matches random batching, saving $68$–$74\%$ of output tokens without losing accuracy relative to solving each task individually. On the remaining five datasets, where random batching degrades accuracy, fragility-aware batching preserves it while still saving roughly $32\%$.\looseness=-1

These results show that much of the thinking an LLM spends on a task in isolation is not needed to solve it. Specifically, on GPQA-Diamond, GSM8K and MMLU-Redux, grouping tasks by fragility saves a substantial share of tokens at no cost in accuracy.\looseness=-1

\subsection{Predicting fragility on held-out tasks}
\label{sec:predict-fragility}

Our results show that the set of fragile tasks is stable for a given model, which hints at the existence of steady properties that define fragility. If such properties exist, fragility should be predictable for tasks the model has never seen under composition. Hence, we extend the intervention to unseen tasks by predicting their fragility with a simple linear predictor.
We encode each task description with Qwen3 Embedding 8B~\citep{zhang2025qwen3embeddingadvancingtext}, keep the embeddings fixed, and train a separate linear predictor for each evaluated language model. To generate labels, we evaluate each task in the training split with 10 trials at every composition size $k \in \{3, 6, 10, 20\}$, placing it in a different composition each time. Then, for each composition size $k$ the predictor outputs the probability that the task performs at least as well in composition as in isolation. To obtain predictions on unseen tasks, we use $5$-fold cross-fitting~\citep{chernozhukov2024doubledebiasedmachinelearningtreatment}, so that the held-out folds together cover the entire dataset (details in Appendix~\ref{app:probe}). Finally, we feed the predicted accuracy drops into the same batching algorithm as in Section~\ref{sec:oracle-fix}.\looseness=-1

Table~\ref{tab:batching_results} reports the results of the linear predictor. Guided by the predictor, fragility-aware batching limits the accuracy loss to at most $0.5$\,pp on every dataset and saves $31.0\%$ of tokens on average. Because few training examples are available, the predictor is simple, and it works best on tasks with short, plain descriptions such as Integrals, GSM8K and OneWordQA. This simple demonstration thus shows that fragility is partly predictable from task properties encoded in the embeddings. Scaling such predictors further could enable saving tokens on real-world unseen tasks, just by batching them well.\looseness=-1

\Needspace{4\baselineskip}
\noindent\textbf{Fragility-aware batching on a frontier model.} We further test whether fragility-aware batching carries over to flagship frontier LLMs. Since the intervention requires neither fine-tuning nor access to internal states and works purely by restructuring compositions, we can apply it directly to Claude Opus 5 on GPQA-Diamond (Table~\ref{tab:batching_results}). In isolation, Claude Opus 5 reaches an accuracy of $89.5\%$ with 689 output tokens per task. Random composition with $k{=}6$ lowers accuracy by $0.9$\,pp and saves $18\%$ of output tokens. Batching by measured fragility does better on both metrics, losing just $0.4$\,pp while saving $24\%$ of generated tokens. The linear predictor, trained on fewer trials for Opus than for the open-weight models, transfers less well: it saves almost no tokens and loses $2.0$\,pp. At least in its consequences, the mechanism we found in open-weight models carries over to a frontier model.\looseness=-1

\section{Effects of post-training on compositional degradation}

\begin{figure}[!htbp]
  \centering
  \includegraphics[width=0.49\linewidth, valign=t]{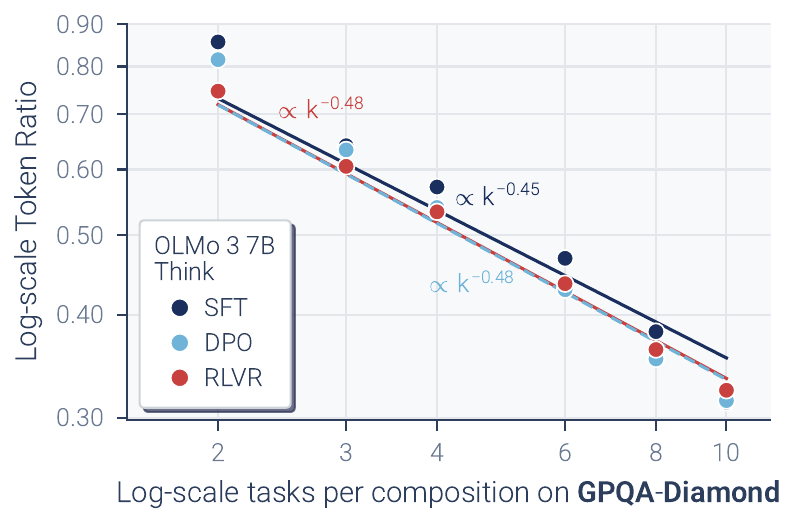}\hfill
  \includegraphics[width=0.49\linewidth, valign=t]{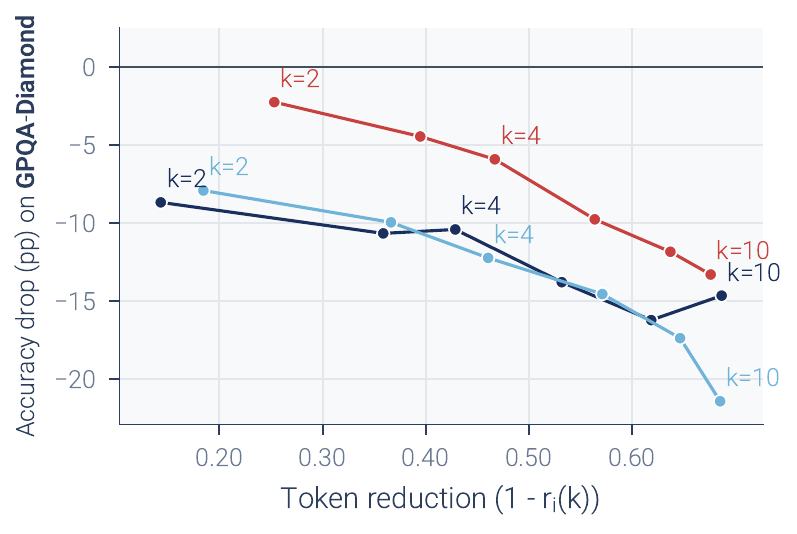}
  \caption{Effects of implicit budgeting across three post-training stages of OLMo~3~7B on GPQA-Diamond. \textbf{Left:} Token ratio $r_i(k) = T_i(k)/T_i(1)$ as a function of composition size $k$. \textbf{Right:} Accuracy drop as a function of the token ratio.}
  \label{fig:olmo3-implicit-budgeting-post-training}
\end{figure}

The models in our analysis degrade very differently under composition, most clearly between the two model families, which differ in their post-training strategy. This hints that post-training strongly influences how well a model handles compositions. To investigate this, we repeat our analysis on OLMo~3~7B \citep{olmo2026olmo3} at three stages of its training curriculum: SFT \citep{ouyang2022traininglanguagemodelsfollow}, DPO \citep{rafailov2024directpreferenceoptimizationlanguage} and RLVR \citep{lambert2025tulu3pushingfrontiers,deepseekai2025deepseekr1incentivizingreasoningcapability}.\looseness=-1

Figure~\ref{fig:olmo3-implicit-budgeting-post-training} summarizes the results on OLMo~3~7B. All three stages show very similar implicit budgeting under composition, and their token counts shrink at similar rates. Nevertheless, the same reduction affects accuracy differently. At similar token ratios, the final RLVR stage loses substantially less accuracy than the SFT and DPO stages. These degrade by similar amounts and also tend to fail on the same fragile tasks. Correlating their accuracy degradation on GPQA-Diamond, MATH-500 and MMLU-Redux as in Section~\ref{sec:transfer-fragility-set} gives a disattenuated Pearson $\rho$ of $1.03\pm0.35$. RLVR's fragile set tends to overlap less with theirs, with $\rho = 0.69\pm0.23$ against SFT and $\rho = 0.89\pm0.26$ against DPO.\looseness=-1

We conclude that RLVR post-training mitigates the accuracy drop incurred by implicit budgeting under composition, but it does not eliminate the misallocation of the thinking budget in the first place, which remained similar across the stages.\looseness=-1

\section{Related work}
\label{sec:relatedworks}

\noindent
Previous work has examined both how task composition affects accuracy and reasoning length and how batching can improve efficiency or reliability. Several studies ask whether models can solve tasks as reliably together as they do separately~\citep{pan2025reststresstestinglarge,wang2025zempe,chen2026multiinstance}. Related studies examine whether failures to meet multiple response requirements can be explained by independent errors~\citep{harada2025instructions,vasileva2026large}. These studies document accuracy losses under composition, although their causes are not yet fully understood.
At the same time, batching methods have shown promise for improving efficiency and reliability~\citep{xu2026robatch,ji2025obp}, with accuracy gains reported in some settings~\citep{srivastava2026constraint,yang2026batchofthought,lin2024batchpromptaccomplish} (consistent with our observations on light-reasoning datasets OneWordQA and StrategyQA, see Appendix~\ref{app:batching-full}). Batching has also been observed to reduce the number of reasoning tokens used per task~\citep{srivastava2026constraint,deng2026conpress}. The effects of added context also vary across questions. For factual questions, models tend to resist misleading information more successfully when they can also answer related questions correctly, suggesting that accuracy in isolation does not fully capture robustness~\citep{xu2026illusions}.
The number of steps needed to solve a task also matters. Tasks requiring more steps show larger accuracy drops when irrelevant information is added~\citep{yang2025distracted}.
More broadly, researchers distinguish task difficulty from sensitivity to prompt changes~\citep{romanou2026brittlebench} and examine how that sensitivity varies with model size, training and architecture~\citep{qin2026interactions}. Overall accuracy can remain largely unchanged even when irrelevant context changes which questions a model answers correctly~\citep{zhang2026illusion}. Estimates of this sensitivity also depend on how answers are graded. Rigid matching rules can treat differently worded correct answers as errors, exaggerating the apparent sensitivity~\citep{hua2025artifact}.
Alongside these studies, work on reasoning allocation has shown that adding other tasks or irrelevant text can shorten reasoning traces and reduce self-checking~\citep{rodionov2026reasoningshift}. Researchers have also trained models to solve groups of tasks more efficiently under a shared token limit~\citep{yang2026bcr}. However, studies of budget allocation find that models often devote more reasoning to earlier questions than to those worth more points~\citep{fan2026ration}. In a single-task setting, model confidence has also been used to guide a search for shorter reasoning paths~\citep{huang2026sage}.\looseness=-1

These findings can be difficult to reconcile within a common explanation, given the differences in experimental settings, interventions and datasets. Our work offers a unifying perspective on several of these observations. Budget misallocation is consistent with the greater sensitivity of deeper reasoning tasks to irrelevant context~\citep{yang2025distracted}, self-compression~\citep{deng2026conpress}, composition errors that cannot be explained by independent error rates alone~\citep{harada2025instructions}, and the token savings achieved through batching~\citep{srivastava2026constraint, cheng2023batchpromptingefficientinference}. We identify reasoning trace length in isolation as a key variable that helps connect these findings and informs simple, effective interventions. Using this information, we are able to show how to mitigate degradation under composition, even in frontier models.\looseness=-1

\section{Discussion and conclusions}
\label{sec:conclusion}

In this work, we show that language models fail on tasks they solve reliably in isolation once those tasks are combined in a single prompt. The resulting accuracy drop is larger than expected from isolated error rates alone. As more tasks are combined, models use fewer reasoning tokens per task, following an approximate power-law decay with composition size, and the largest failures occur on tasks that require longer reasoning in isolation. This reveals an implicit per-task budget that acts as a double-edged sword. Under composition, fragile tasks receive too little reasoning and fail. In isolation, non-fragile tasks receive more reasoning than they need and waste compute. By batching tasks according to their fragility, however, the implicit budgeting can be manipulated to reduce token usage by about 43\% on average while preserving accuracy, and since it does not require internal states of the model, it also applies to proprietary models such as Claude Opus~5.\looseness=-1

\Needspace{4\baselineskip}
\noindent\textbf{Previous context can cause degradation too.} We deliberately compose independent tasks of the same type, isolating the budget effect in its simplest form. An important question is how this effect extends to richer contexts. Prior work finds that reasoning traces also shorten when the same task is embedded in irrelevant context, conversation history, or a larger task~\citep{rodionov2026reasoningshift}, suggesting that implicit thinking budgeting may extend beyond same-type composition.\looseness=-1

\Needspace{4\baselineskip}
\noindent\textbf{Long-running agents in production.} This is particularly relevant to agentic workflows, where tool outputs, interaction histories, and intermediate results continually accumulate. Agents are widely observed to degrade as their context grows, a phenomenon known as context rot~\citep{hong2025context}. Our results point to a degradation that sets in long before the context fills up. In an agentic workflow, this means a high-effort step can be unknowingly misbudgeted, thus possibly leading to error propagation in subsequent tasks. As a direct practical intervention, agents can keep their context lean and hand reasoning-intensive steps to subagents that start from a fresh context.\looseness=-1

\Needspace{4\baselineskip}
\noindent\textbf{Post-training as a potential mitigation.} Our results suggest that reasoning allocation contributes to the difficulty of composed tasks, even when the constituent tasks are solvable in isolation. Post-training may therefore help mitigate degradation under composition. For instance, integrating budget management into RL pipelines through explicit training on composed tasks could help models allocate reasoning more effectively across a shared prompt.\looseness=-1

\Needspace{4\baselineskip}
\noindent\textbf{Limitations.}
While we show that composition systematically reduces per-task reasoning and disproportionately harms fragile tasks, we do not yet know what drives this reduction. Our study is limited to independent, same-dataset compositions, and detailed reasoning-length analysis is restricted to open-weight models. Extending these results to mixed-domain, dependent, and proprietary-model settings remains important future work.\looseness=-1

\enlargethispage{\baselineskip}
\section*{Acknowledgments}
This project was funded by NSF award CCF-2112665 (TILOS) and an Alexander von Humboldt professorship. Sharut Gupta was supported by a MathWorks Engineering Fellowship. Thies Brockmöller and Chenpan Li are supported by the scholarship of the Studienstiftung des deutschen Volkes, and Chenpan Li is also supported by the scholarship of the Thomas-Weiland Wissenschaftsstiftung. \looseness=-1
\newpage

\bibliographystyle{abbrvnat}
\bibliography{iclr2027_conference}

\clearpage
\appendix
\setlength{\parskip}{6pt plus 1pt minus 1pt}
\section{Appendix}
\label{sec:appendix}

\subsection{Further results}

\subsubsection{Compositional degradation is observable over a wide range of datasets and models}
\label{app:success-conditioned-analysis}

\begin{figure}[!htbp]
  \centering
  \includegraphics[width=0.88\linewidth, valign=t]{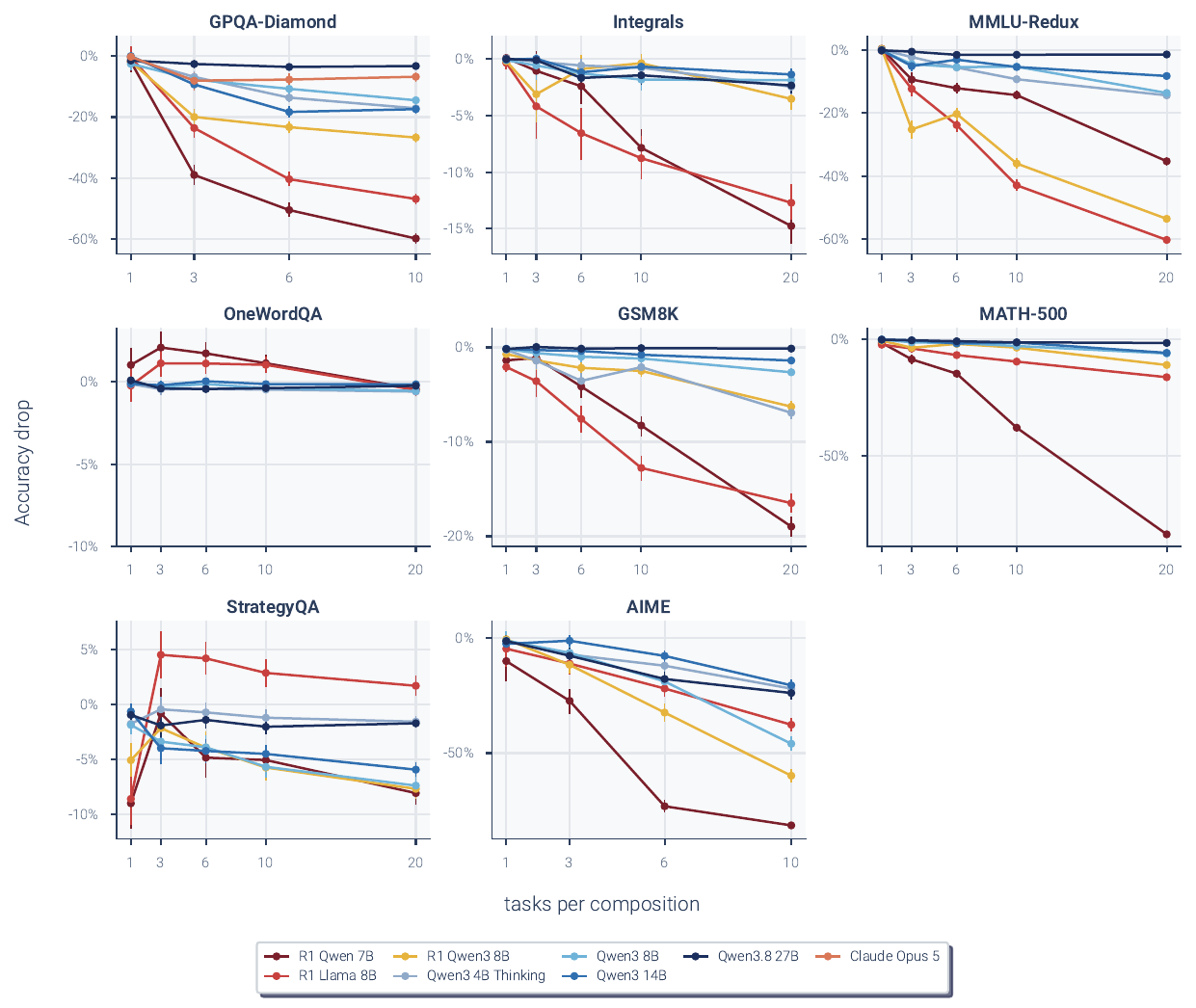}
  \caption{Accuracy drop as a function of composition size $k$ for all datasets and models. All tasks were solved in three out of three isolated attempts, yet accuracy drops once they are composed, on every dataset except OneWordQA and StrategyQA, whose tasks require little reasoning. The R1-distilled models degrade the most, and the extent of degradation varies strongly between datasets.}
\end{figure}

\paragraph{Compositional degradation is not caused by context length or parsing.}  Since composed tasks require more context and use a different answer format than isolated tasks, two natural confounders of compositional degradation are truncation due to context overflow and failures in parsing the composed answer. The results in Table~\ref{tab:context-parsing} show that neither factor accounts for the observed degradation. Context length poses little difficulty: in most datasets, fewer than 1\% of responses reach the context limit, even at the largest $k$ we evaluate. Parsing failures are likewise rare across all datasets except AIME. There, the majority of these failures arise because the model returns fewer answers than there are tasks, which itself constitutes a failure to solve the composition.\looseness=-1

\FloatBarrier
\subsubsection{Composition size influences token ratio}
\label{app:composition-size-token-ratio}
\begin{figure}[!htbp]
  \centering
  \includegraphics[width=0.92\linewidth, valign=t]{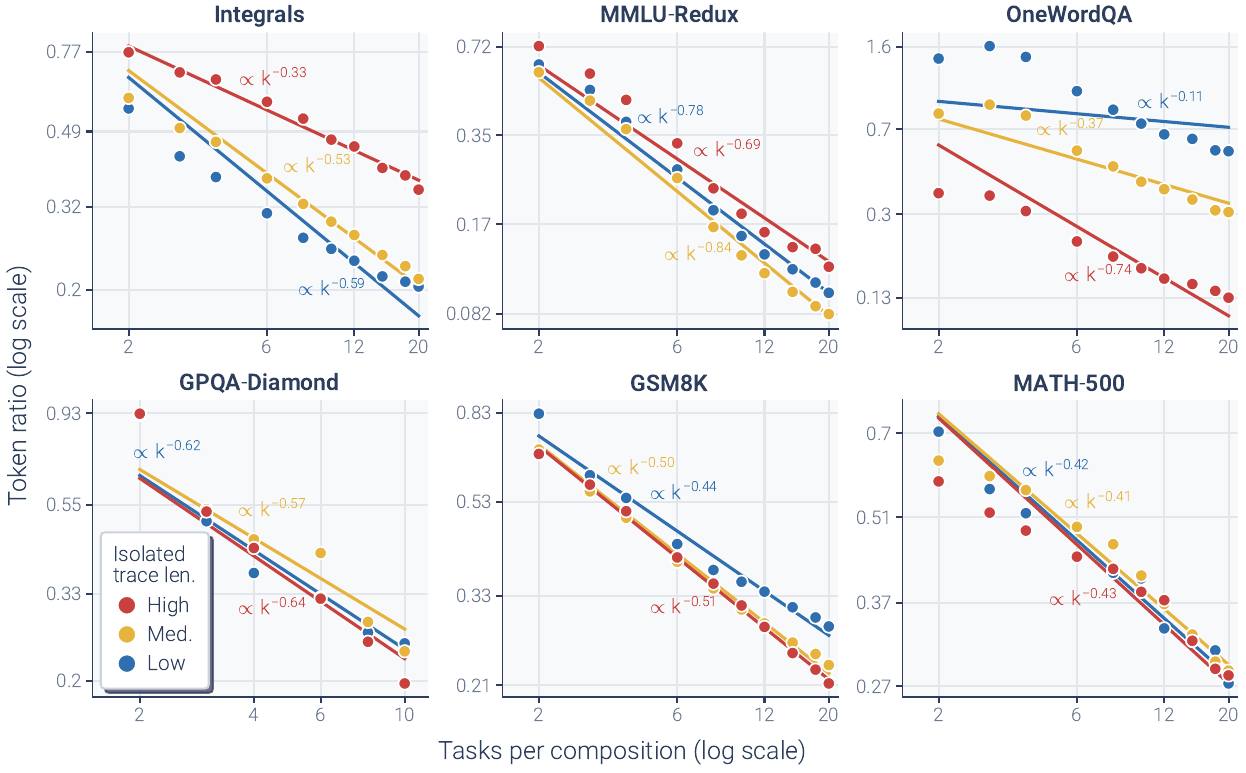}
  \caption{Token ratio on Qwen3 4B Thinking. Power-law fits through $(1, 1)$, one per dataset: median slope $m = -0.46$ (range $[-0.79, -0.34]$), mean $R^2 = 0.94$.}
  \label{fig:ratio-qwen3-4b}
\end{figure}

\begin{figure}[!htbp]
  \centering
  \includegraphics[width=0.92\linewidth, valign=t]{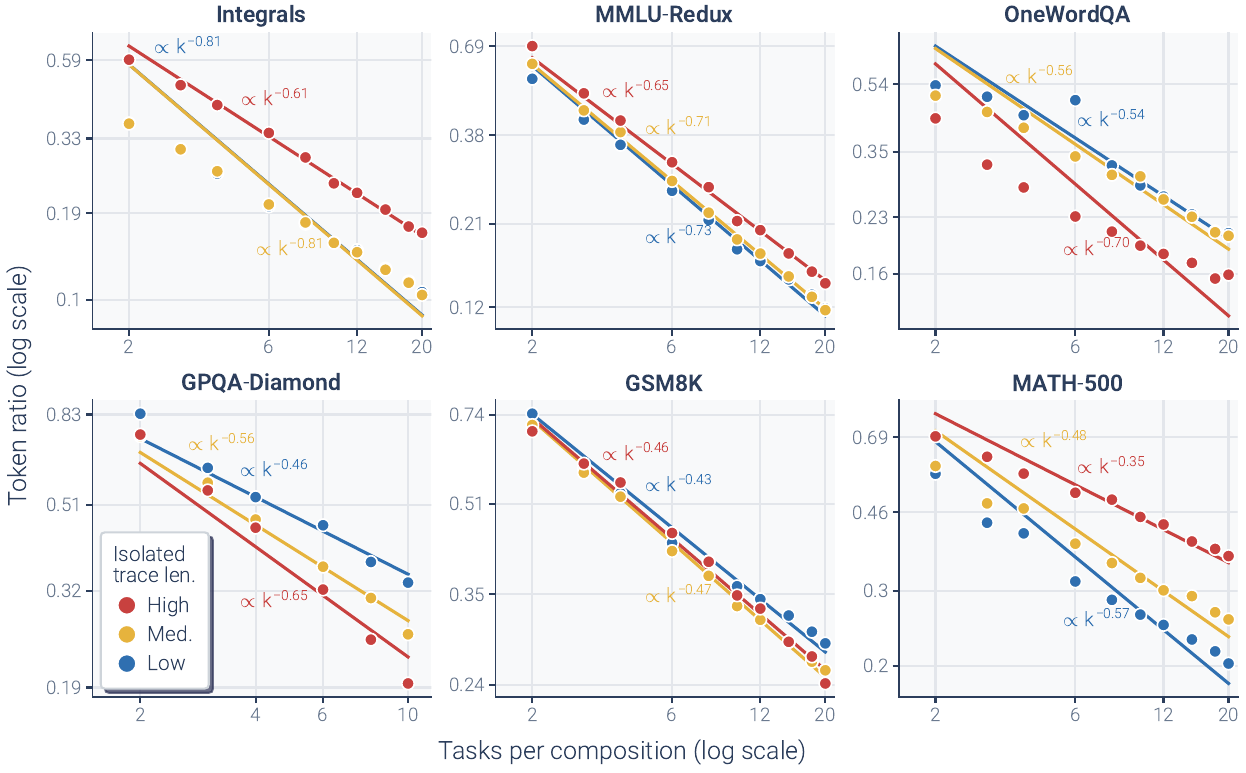}
  \caption{Token ratio on Qwen3 8B. Power-law fits through $(1, 1)$, one per dataset: median slope $m = -0.56$ (range $[-0.72, -0.45]$), mean $R^2 = 0.94$.}
  \label{fig:ratio-qwen3-8b}
\end{figure}

\begin{figure}[!htbp]
  \centering
  \includegraphics[width=0.92\linewidth, valign=t]{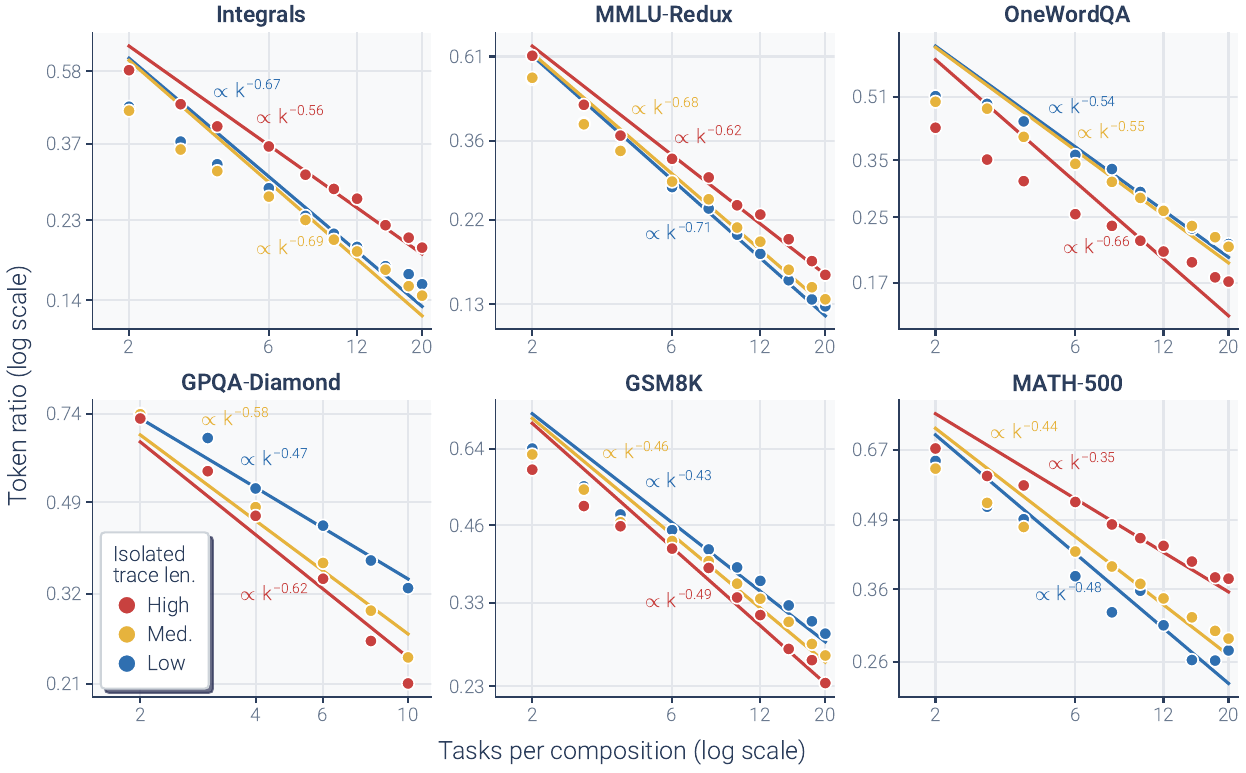}
  \caption{Token ratio on Qwen3 14B. Power-law fits through $(1, 1)$, one per dataset: median slope $m = -0.55$ (range $[-0.73, -0.42]$), mean $R^2 = 0.87$.}
  \label{fig:ratio-qwen3-14b}
\end{figure}

\begin{figure}[!htbp]
  \centering
  \includegraphics[width=0.92\linewidth, valign=t]{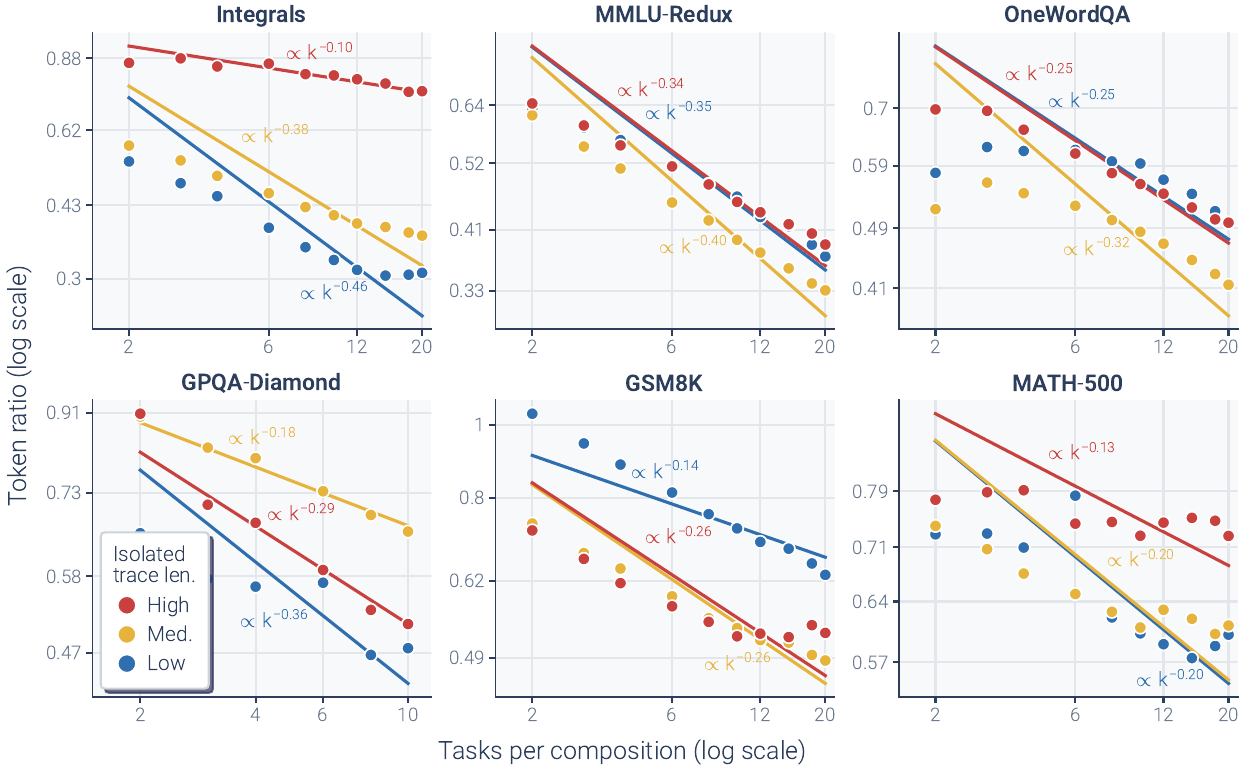}
  \caption{Token ratio on Qwen3.8 27B. Power-law fits through $(1, 1)$, one per dataset: median slope $m = -0.15$ (range $[-0.26, -0.07]$), mean $R^2 = 0.72$.}
  \label{fig:ratio-qwen3-8-27b}
\end{figure}

\begin{figure}[!htbp]
  \centering
  \includegraphics[width=0.92\linewidth, valign=t]{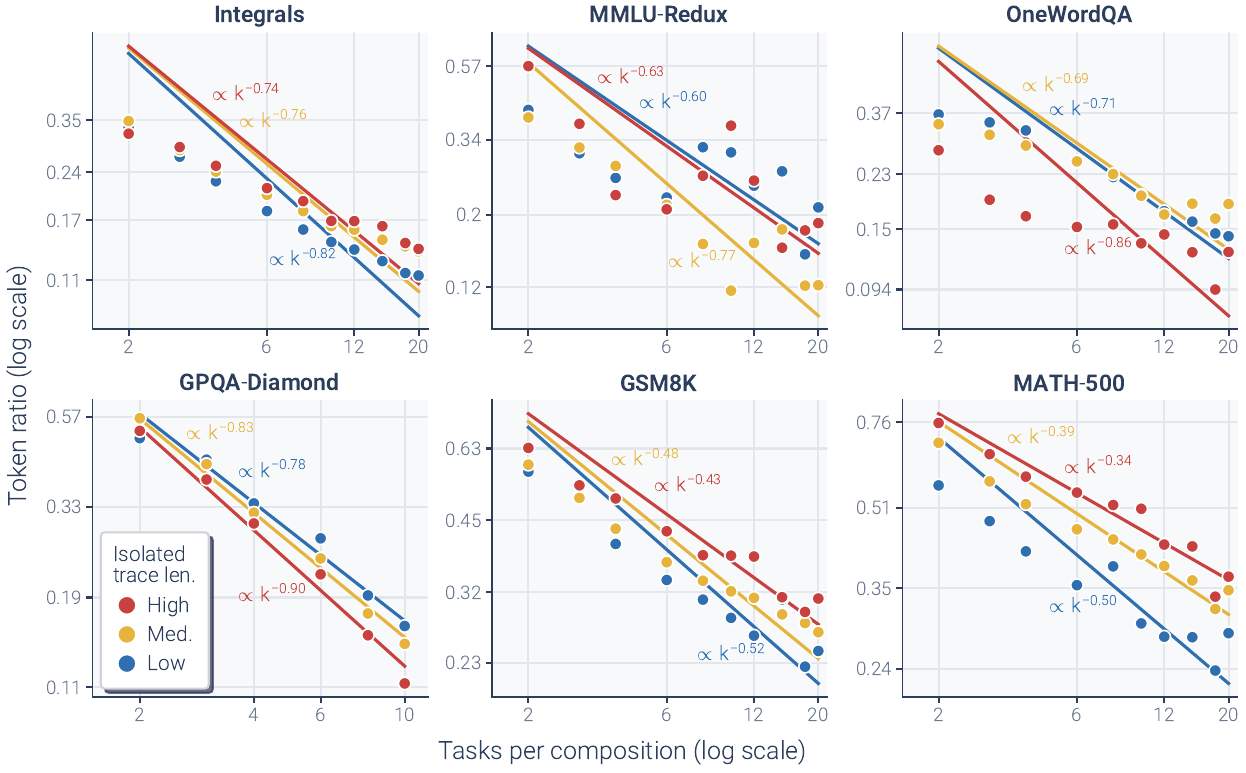}
  \caption{Token ratio on R1 Llama 8B. Power-law fits through $(1, 1)$, one per dataset: median slope $m = -0.68$ (range $[-0.78, -0.35]$), mean $R^2 = 0.59$.}
  \label{fig:ratio-r1-llama-distill-8b}
\end{figure}

\begin{figure}[!htbp]
  \centering
  \includegraphics[width=0.92\linewidth, valign=t]{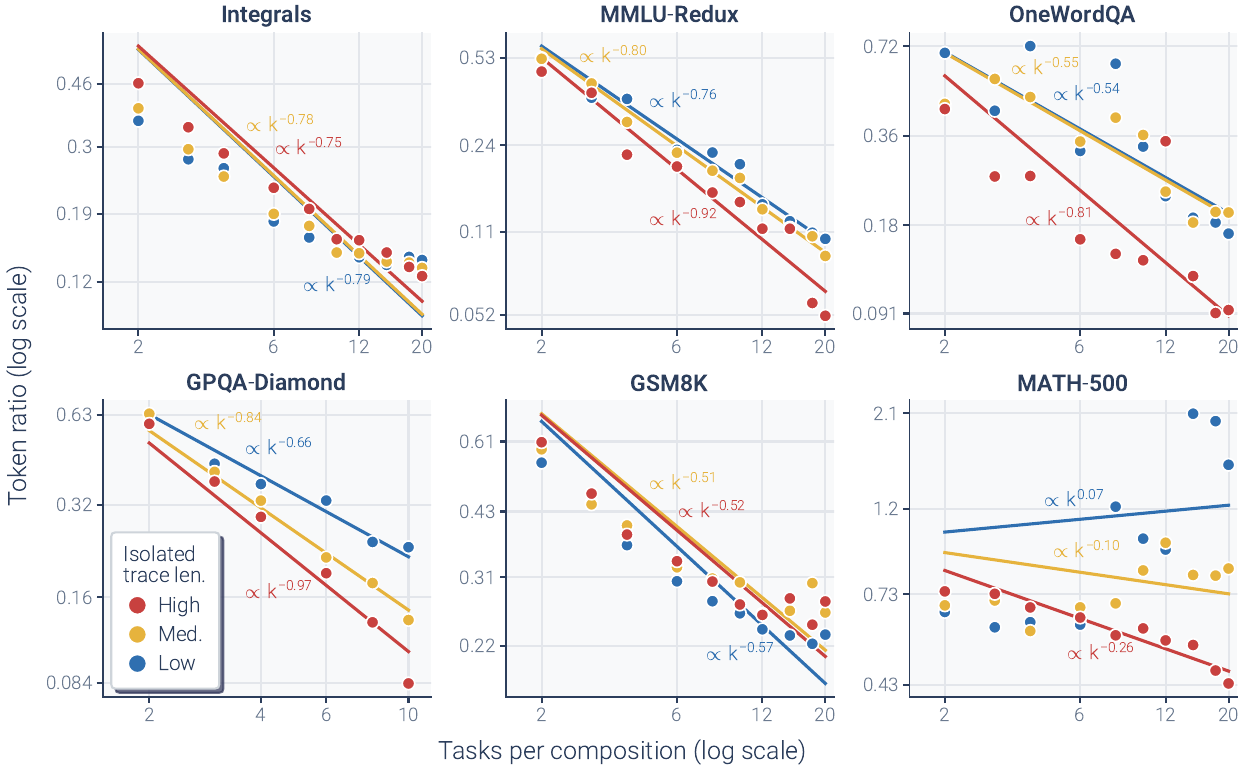}
  \caption{Token ratio on R1 Qwen 7B. Power-law fits through $(1, 1)$, one per dataset: median slope $m = -0.62$ (range $[-0.78, -0.04]$), mean $R^2 = 0.62$.}
  \label{fig:ratio-r1-distill-7b}
\end{figure}

\begin{figure}[!htbp]
  \centering
  \includegraphics[width=0.92\linewidth, valign=t]{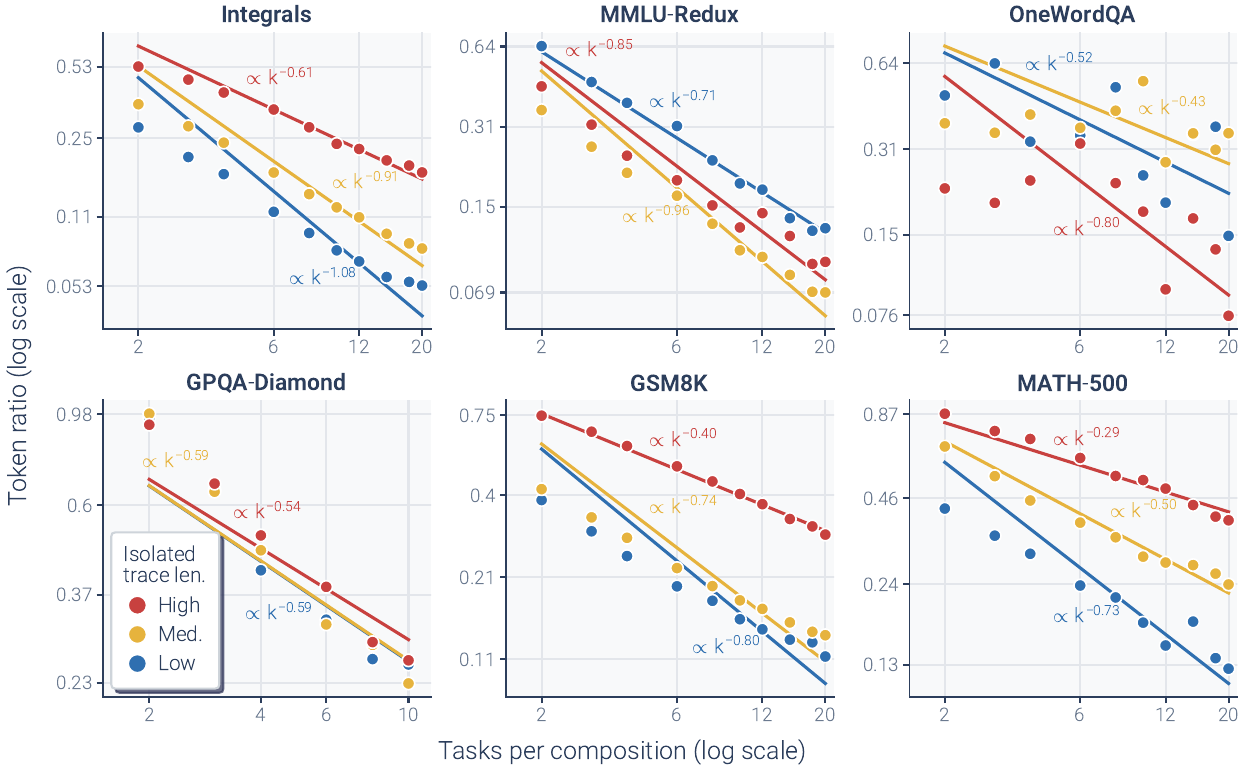}
  \caption{Token ratio on R1 Qwen3 8B. Power-law fits through $(1, 1)$, one per dataset: median slope $m = -0.61$ (range $[-0.88, -0.46]$), mean $R^2 = 0.55$.}
  \label{fig:ratio-r1-qwen3-8b}
\end{figure}

Figures~\ref{fig:ratio-qwen3-4b}--\ref{fig:ratio-r1-qwen3-8b} show the token ratio $r_i(k) = T_i(k)/T_i(1)$ as a function of composition size $k$ for all seven open-weight models. Tasks are binned by the quantile of their isolated trace length into Low ($\leq 20\%$), Medium ($20$--$80\%$), and High ($> 80\%$). On log-log axes, the ratio decreases approximately linearly in $\log k$ for nearly all models and datasets, indicating that LLMs shrink the relative token usage of each task under composition by $k^{m}$. The exception is R1 Qwen 7B on MATH-500, where the ratio of the Low bin rises above 1 and the fit over all tasks is flat ($m = -0.04$, $R^2 = -0.43$). This reduction shows no systematic dependence on $T_i(1)$, since the ordering of the length bins varies across models and datasets and no bin is consistently reduced more than the others. We fix each power-law fit to pass through $(1, 1)$, since a composition of a single task has a token ratio of 1 by construction. Each caption summarizes the six fits of a model, one per dataset over all of its tasks, by the median and range of the slope $m$ and the mean $R^2$.\looseness=-1

\FloatBarrier
\subsubsection{Token ratio influences accuracy degradation}
\label{app:token-ratio-accuracy}

\begin{figure}[!htbp]
  \centering
  \includegraphics[width=0.92\linewidth, valign=t]{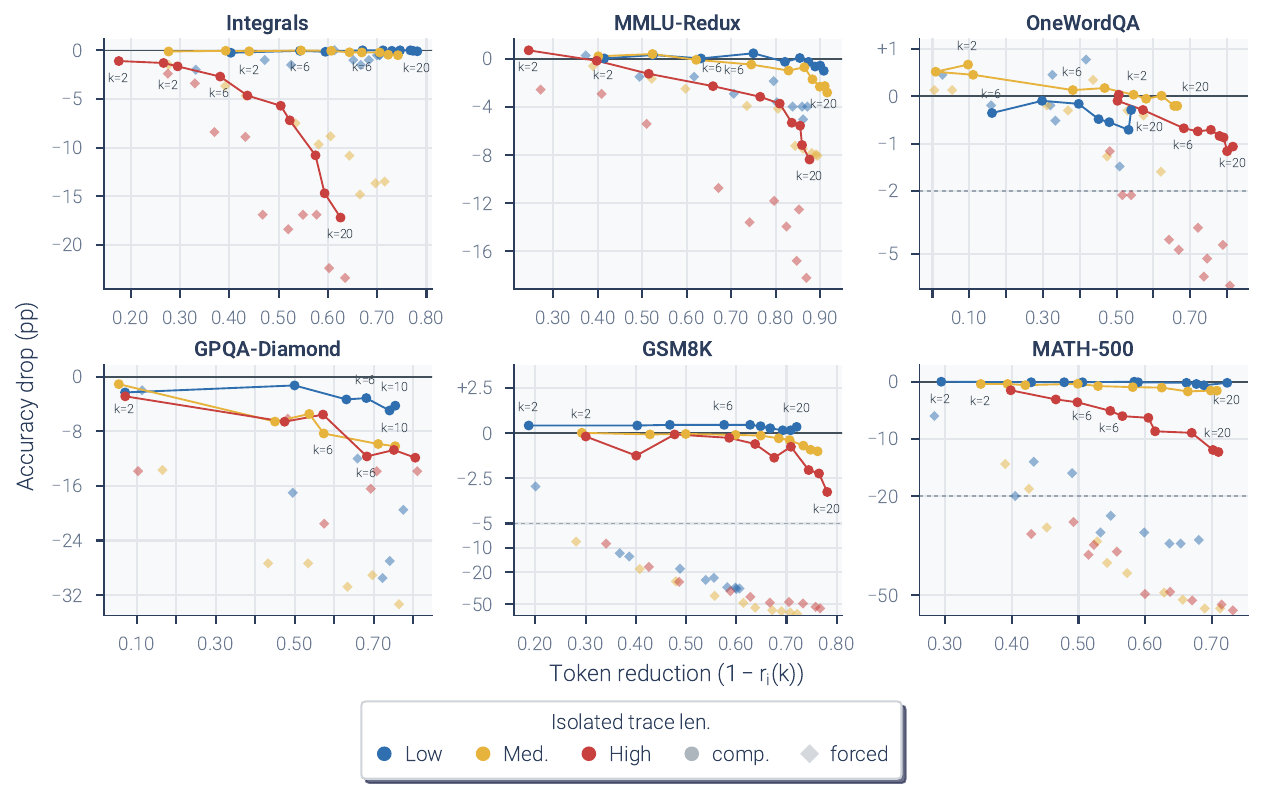}
  \caption{Accuracy drop versus relative thinking token reduction for Qwen3 4B Thinking.}
  \label{fig:accuracy-deg-qwen3-4b-thinking}
\end{figure}

\begin{figure}[!htbp]
  \centering
  \includegraphics[width=0.92\linewidth, valign=t]{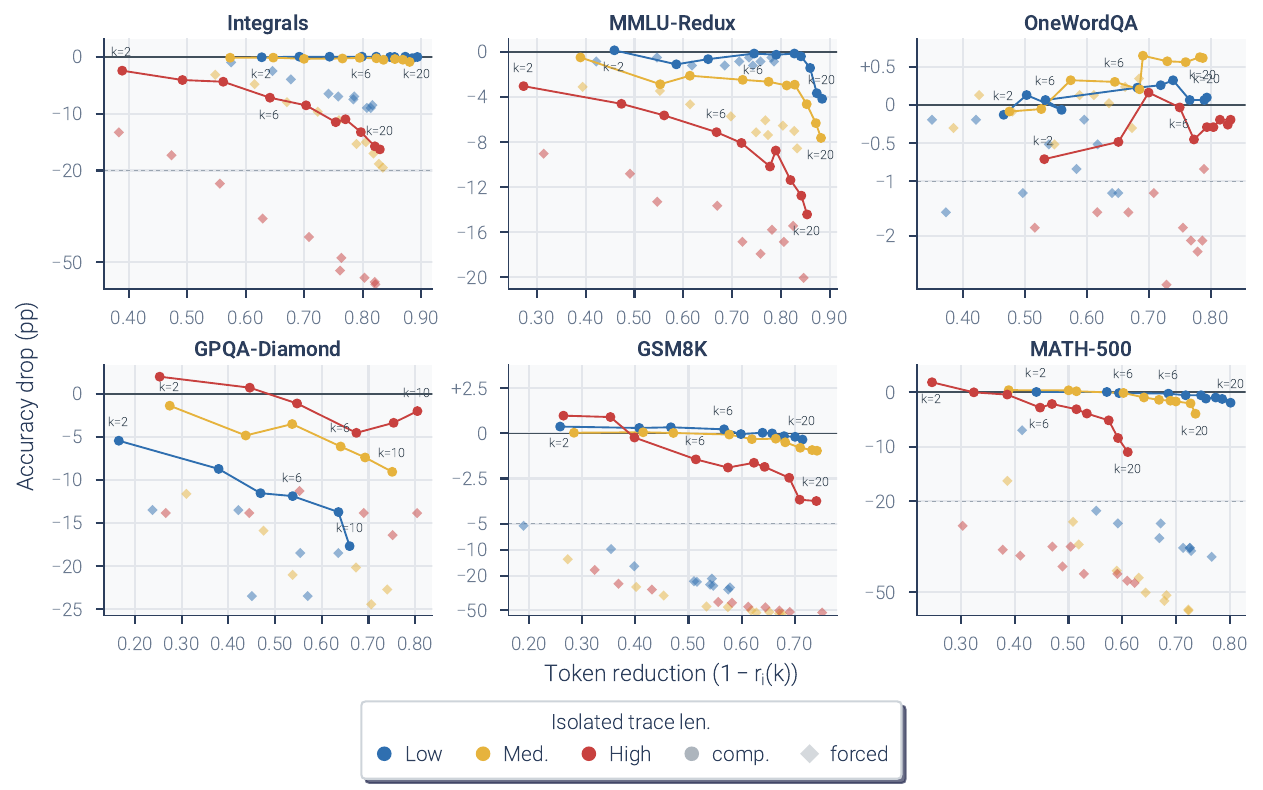}
  \caption{Accuracy drop versus relative thinking token reduction for Qwen3 8B.}
  \label{fig:accuracy-deg-qwen3-8b}
\end{figure}

\begin{figure}[!htbp]
  \centering
  \includegraphics[width=0.92\linewidth, valign=t]{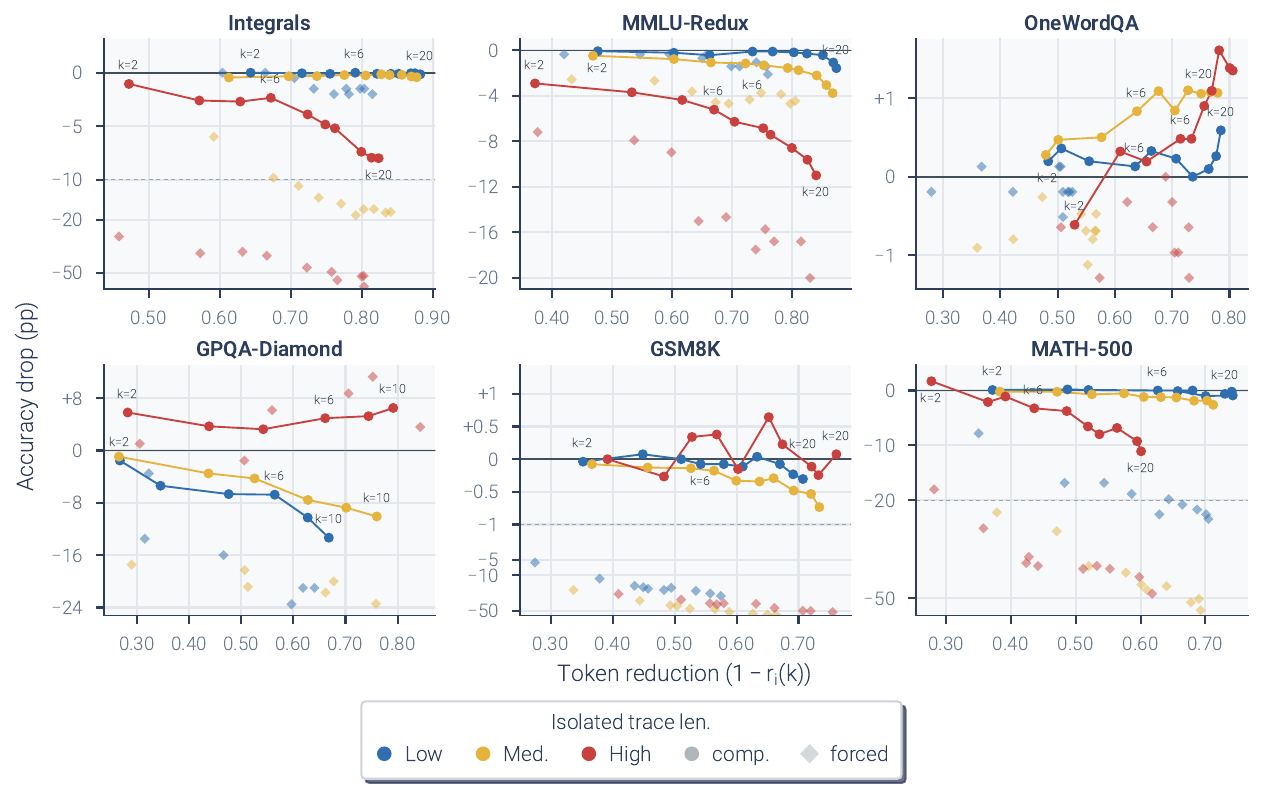}
  \caption{Accuracy drop versus relative thinking token reduction for Qwen3 14B.}
  \label{fig:accuracy-deg-qwen3-14b}
\end{figure}

\begin{figure}[!htbp]
  \centering
  \includegraphics[width=0.92\linewidth, valign=t]{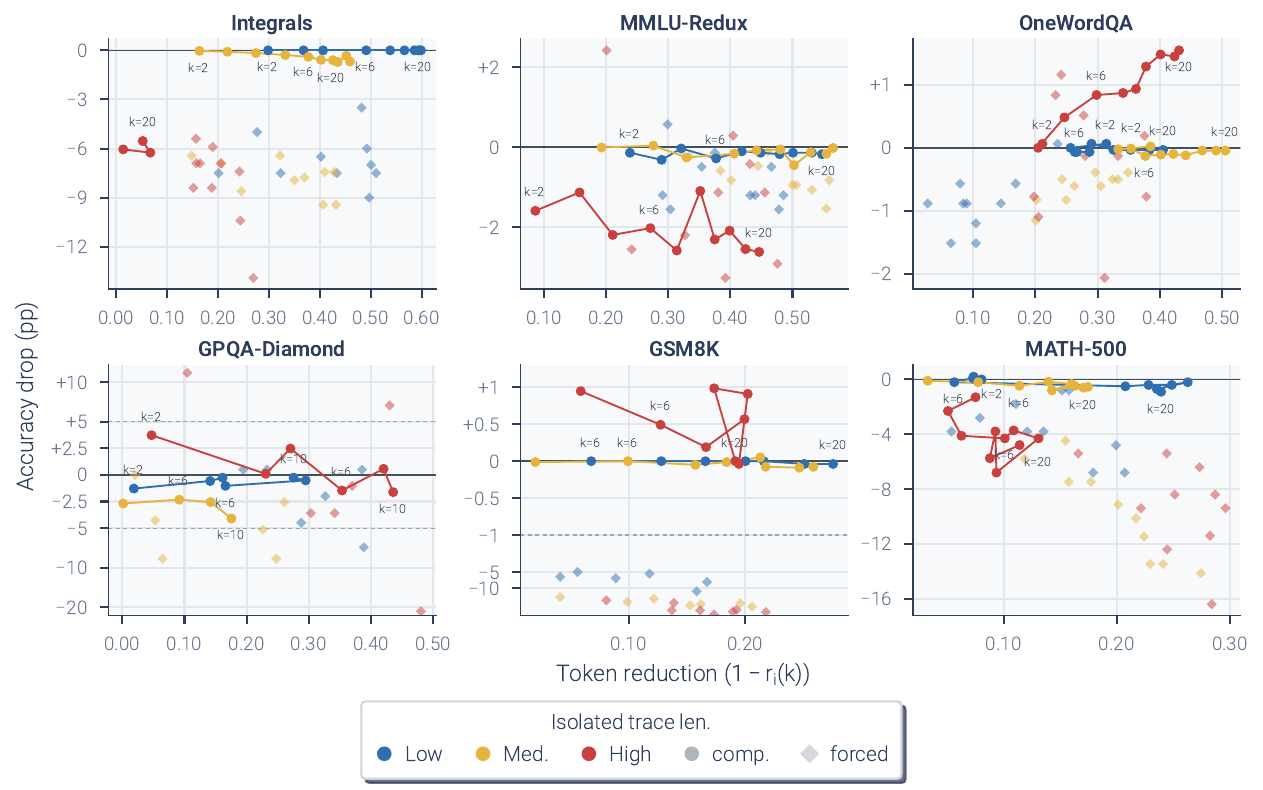}
  \caption{Accuracy drop versus relative thinking token reduction for Qwen3.8 27B.}
  \label{fig:accuracy-deg-qwen3-8-27b}
\end{figure}

\begin{figure}[!htbp]
  \centering
  \includegraphics[width=0.92\linewidth, valign=t]{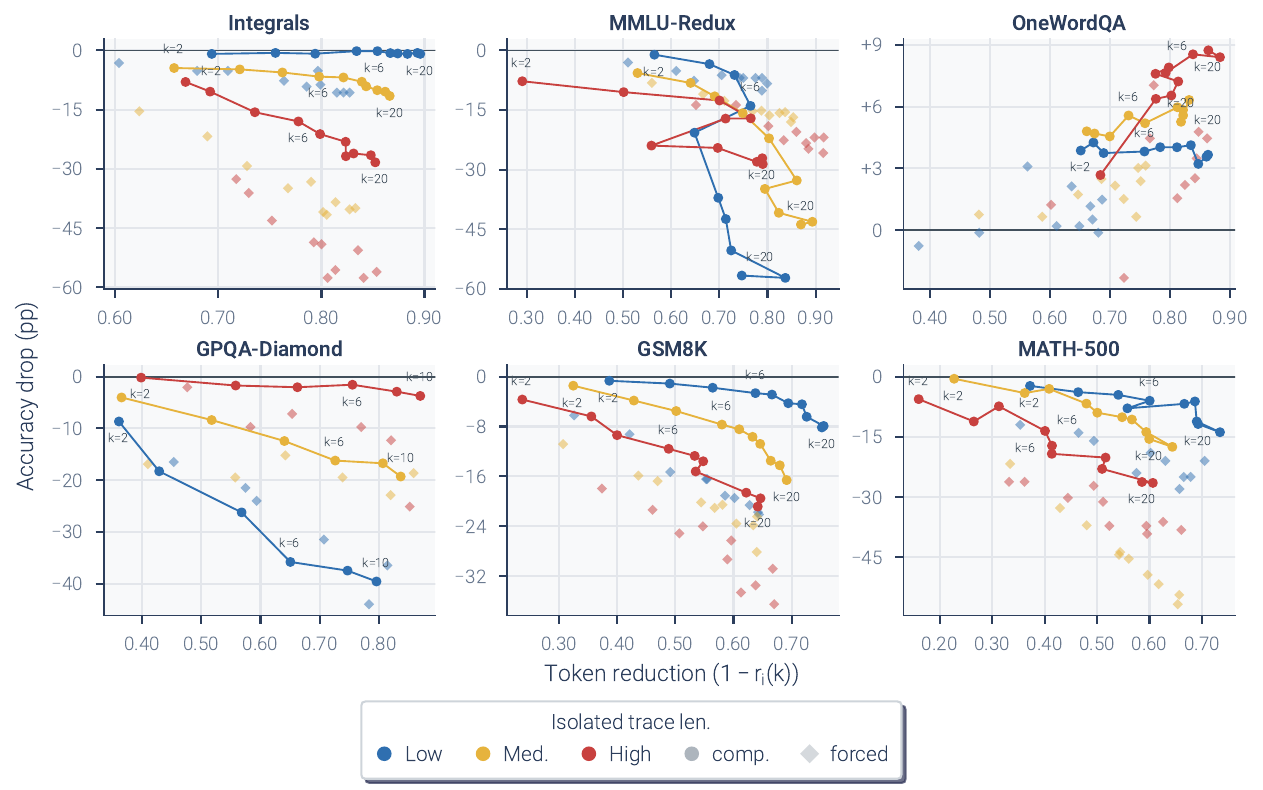}
  \caption{Accuracy drop versus relative thinking token reduction for R1 Llama 8B.}
  \label{fig:accuracy-deg-r1-llama-distill-8b}
\end{figure}

\begin{figure}[!htbp]
  \centering
  \includegraphics[width=0.92\linewidth, valign=t]{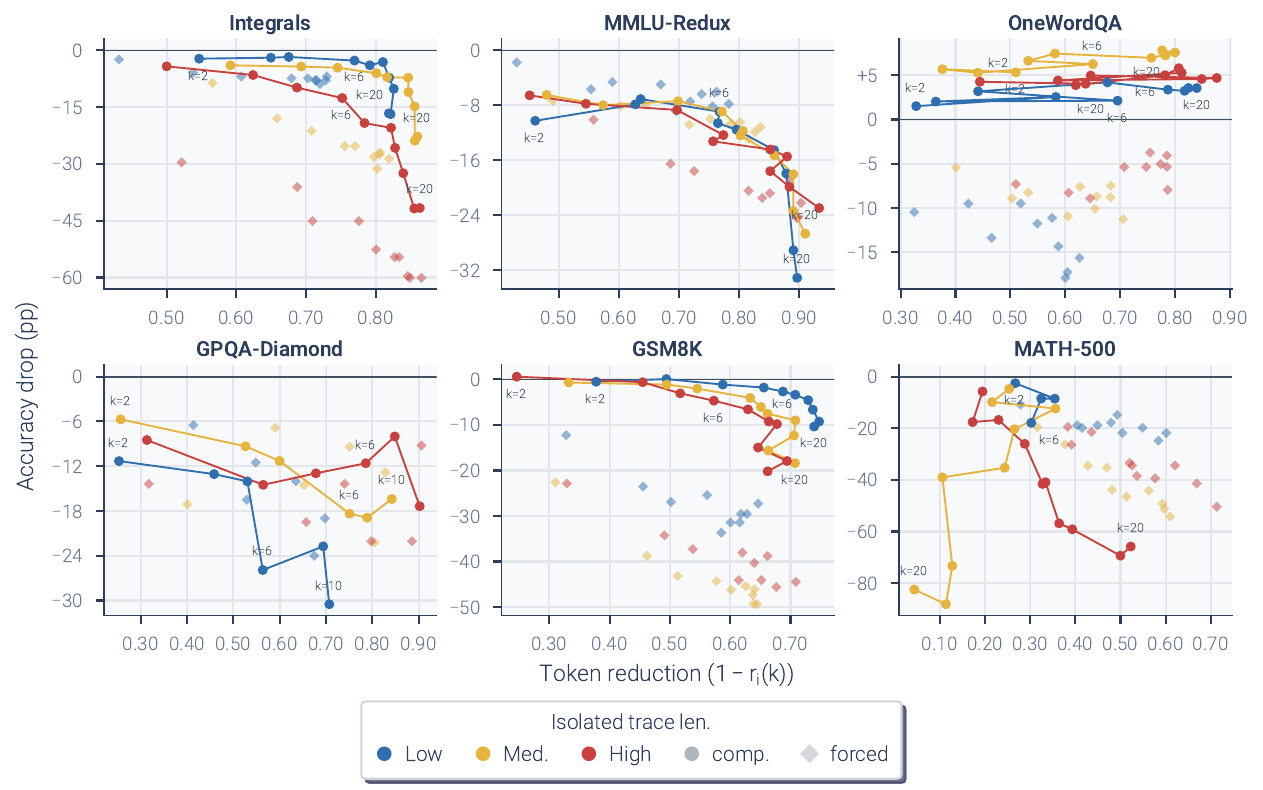}
  \caption{Accuracy drop versus relative thinking token reduction for R1 Qwen 7B.}
  \label{fig:accuracy-deg-r1-qwen-distill-7b}
\end{figure}

\begin{figure}[!htbp]
  \centering
  \includegraphics[width=0.92\linewidth, valign=t]{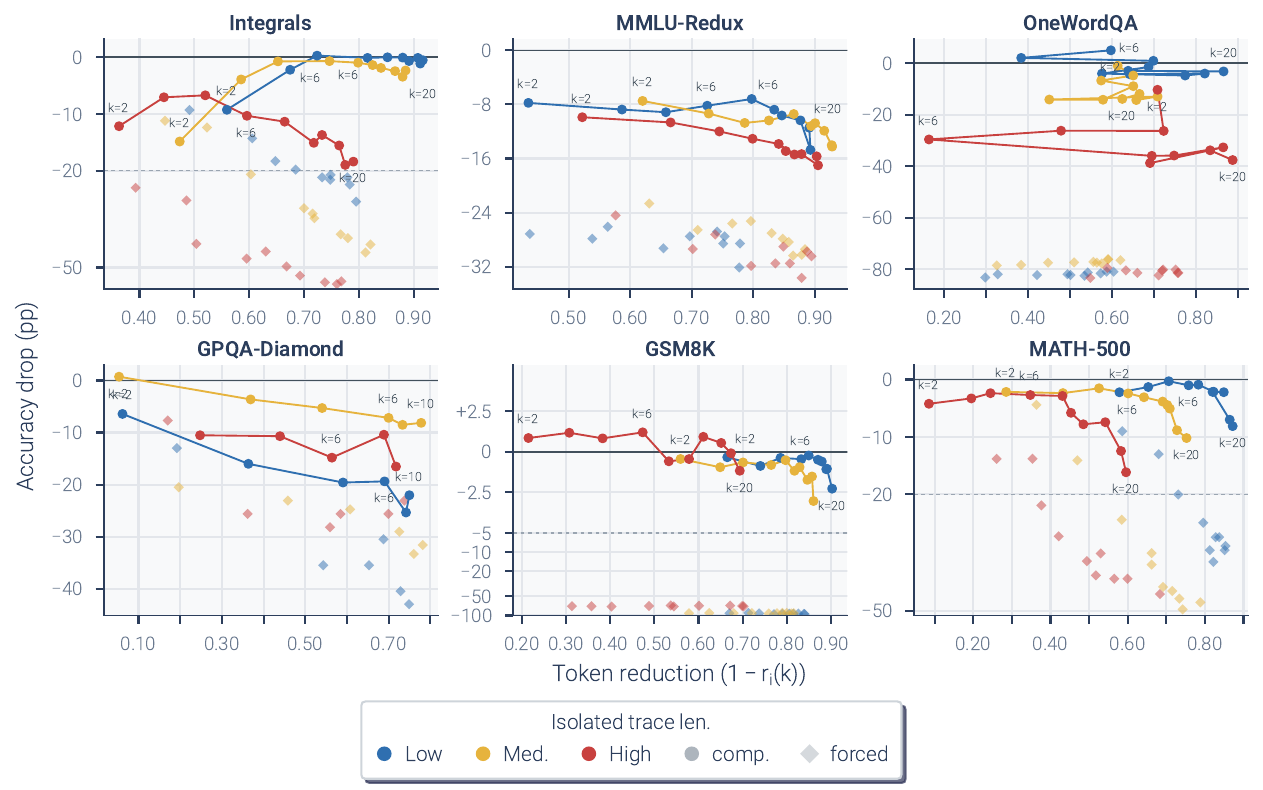}
  \caption{Accuracy drop versus relative thinking token reduction for R1 Qwen3 8B.}
  \label{fig:accuracy-deg-r1-qwen3-8b}
\end{figure}

Figures~\ref{fig:accuracy-deg-qwen3-4b-thinking}--\ref{fig:accuracy-deg-r1-qwen3-8b} extend the analysis of Figure~\ref{fig:token-ratio-vs-accuracy} to further models and datasets. We restrict this analysis to a subset of six datasets, chosen to span a wide range of tasks from math, logical inference and linguistics, as running it on all datasets would be prohibitively expensive.\looseness=-1

In each plot, tasks are binned by the quantile of their isolated thinking length $T_i(1)$ into Low ($\leq 20\%$), Medium ($20$--$80\%$), and High ($> 80\%$), and the accuracy drop is shown as a function of the relative thinking token reduction. Below the dashed line, the y-axis is logarithmic. The pattern holds across all models. Accuracy drops as the reduction grows, and in most cases tasks with long isolated thinking traces degrade substantially more than tasks with short ones.\looseness=-1

\FloatBarrier
\paragraph{Budget forcing causes a degradation analogous to composition.}
To rule out possible context effects, we take the budget a task receives when it is solved in composition and apply it to the same task solved in isolation. Following \citet{muennighoff2025s1simpletesttimescaling}, we inject \texttt{</think>} once the thinking trace reaches this budget. In composition, the model also spends tokens on restating the question and checking the output format. To cover this constant overhead, we add the full share of unattributed tokens (tokens not attributed to any task) from the composed solution to the budget. Figures~\ref{fig:token-ratio-vs-accuracy} and \ref{fig:accuracy-deg-qwen3-4b-thinking}--\ref{fig:accuracy-deg-r1-qwen3-8b} show the results as ``forced''. We find that stopping the thinking abruptly causes substantially larger degradation than the planned reduction of thinking tokens under composition. Still, the same pattern emerges. Degradation rises monotonically as the budget shrinks, and tasks with higher reasoning requirements, measured by their thinking length in isolation, consistently degrade more than tasks with low requirements. In terms of its effect on accuracy, composing tasks therefore appears to act analogously to forcing a budget.\looseness=-1

\FloatBarrier
\subsubsection{Budgeting is slightly position dependent}
\label{app:positional-bias}

\begin{figure}[!htbp]
  \centering
  \includegraphics[width=\linewidth, valign=t]{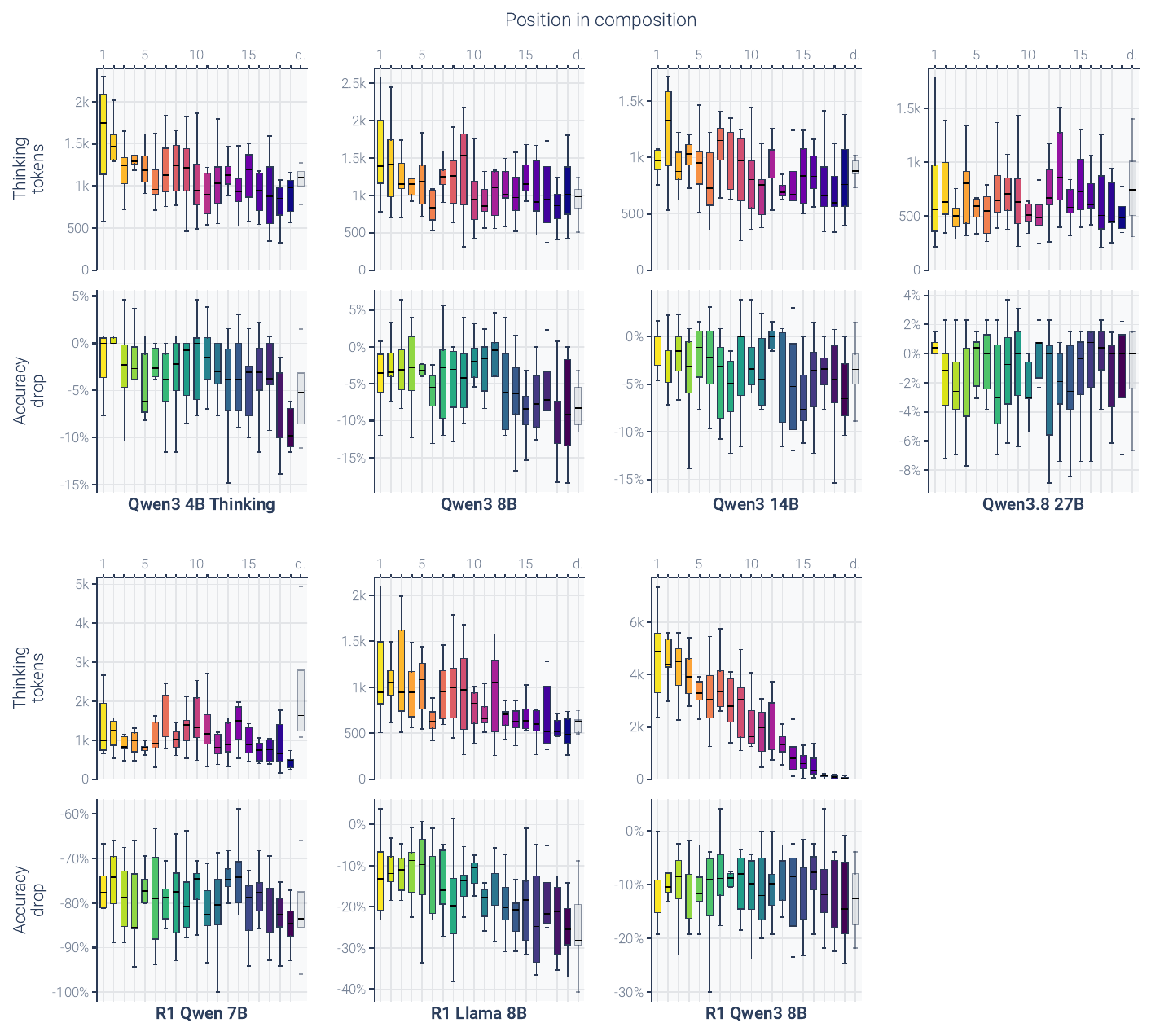}
  \caption{Positional dependence of thinking budget allocation and compositional degradation on MATH-500 at composition size $k=20$. The upper panels show thinking tokens (yellow-red-purple) and the lower panels show the accuracy drop (yellow-green-purple). The last position in each composition holds the dummy task, which we exclude from the analysis because our method cannot reliably determine its token usage (Appendix~\ref{app:trace-segmentation}).}
  \label{fig:position-effect}
\end{figure}

We also report results on how the implicit budgeting depends on position, where a task’s position in a composition is simply its position in the prompt. Previous work has observed a position bias in thinking tokens \citep{pan2025reststresstestinglarge}, so we ask whether this bias also influences the accuracy degradation through composition.\looseness=-1

Figure~\ref{fig:position-effect} plots the budgeting and accuracy drop by position on MATH-500 across all models. We observe a similar budgeting effect as previously described in the literature: models use fewer thinking tokens towards the end of a composition. Remarkably, as a result, the accuracy drop grows towards the end in most models, in line with our earlier observations.\looseness=-1

Qwen3.8~27B does not show a positional bias, and thus its accuracy stays constant across positions, which hints that the training recipe is the main driver of this bias. The original Qwen3~8B and its R1-distilled version point the same way. They share the same architecture, yet R1~Qwen3~8B reduces thinking tokens substantially more at later positions, which again suggests that post-training has a large influence on budgeting behavior. Future work should investigate this phenomenon further, since it can impact performance when high-effort tasks are placed at the end of a prompt.\looseness=-1

\FloatBarrier
\subsubsection{Correlating accuracy drop across models and composition sizes}

\begin{figure}[!htbp]
  \centering
  \includegraphics[width=\linewidth, valign=t]{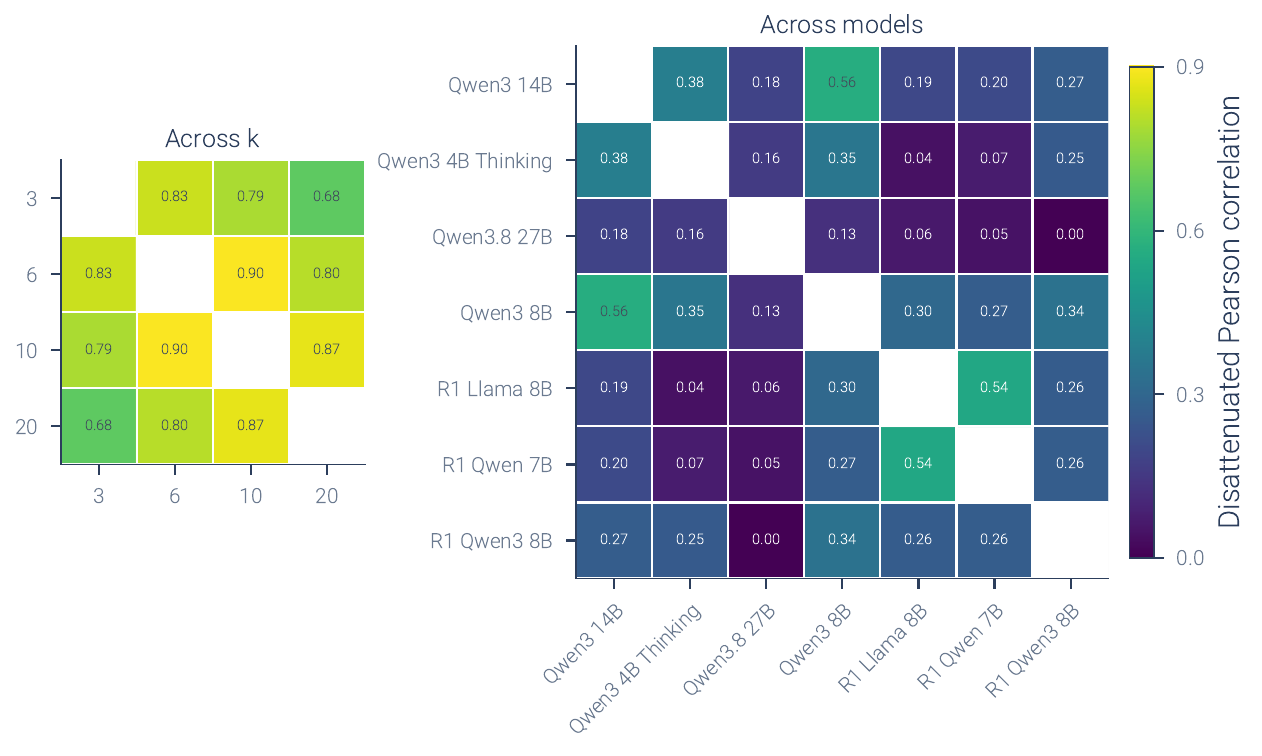}
  \caption{Disattenuated Pearson $r$ correlation of accuracy degradation across different conditions, i.e. model and composition size. Each pair of conditions (a cell in the heatmap) averages over datasets and, on the left, over models, on the right, over composition sizes. \textbf{Left:} The correlation is substantially larger when comparing accuracy degradation within the same model across different composition sizes $k$. \textbf{Right:} By contrast, the correlation between different models is low, especially across model families, suggesting different architectures and training recipes lead to different tasks being fragile.}
  \label{fig:correlation}
\end{figure}

\clearpage
\subsubsection{Fragility-aware batching}
\label{app:batching-full}

\begin{table}[H]
\centering
\renewcommand{\arraystretch}{0.9}
\caption{Task accuracy (Acc.) and output tokens (Tok., includes thinking and answer) per task under different batching policies for all models and datasets. The $k{=}1$ columns report absolute values. All other columns report absolute values with the change with respect to $k{=}1$ in parentheses, as an absolute difference in percentage points for accuracy and as a relative change in \% for tokens. Within each policy, the left column is accuracy and the right column is tokens. The Average row reports the mean over all models shown, eight on GPQA-Diamond and seven on the other datasets: absolute values for $k{=}1$ and the mean change for all other policies. On OneWordQA, the Average leaves out R1 Qwen3 8B (grey), whose answers in composition do not follow our result contract, which leads to increased parsing errors.}
\label{tab:batching_results_full}
\resizebox*{!}{0.79\textheight}{%
\begin{tabular}{lllll|llll}
\toprule
 & \multicolumn{2}{c}{Isolated, $k{=}1$} & \multicolumn{2}{c|}{Random batching, $k{=}6$} & \multicolumn{2}{c}{Fragility-aware (measured)} & \multicolumn{2}{c}{Fragility-aware (predictor)} \\
\cmidrule(lr){2-3} \cmidrule(lr){4-5} \cmidrule(lr){6-7} \cmidrule(lr){8-9}
 & \multicolumn{1}{c}{Acc.} & \multicolumn{1}{c}{Tok.} & \multicolumn{1}{c}{Acc.} & \multicolumn{1}{c|}{Tok.} & \multicolumn{1}{c}{Acc.} & \multicolumn{1}{c}{Tok.} & \multicolumn{1}{c}{Acc.} & \multicolumn{1}{c}{Tok.} \\
\midrule
\multicolumn{9}{l}{\textit{GPQA-Diamond}} \\
\midrule
Qwen3 4B T. & 66.7\% & 7663 & \ppcell{56.4\%}{-10.3} & \relcell{2470}{-67.8} & \ppcell{66.8\%}{+0.2} & \relcell{5200}{-32.1} & \ppcell{67.3\%}{+0.7} & \relcell{7534}{-1.7} \\
Qwen3 8B & 58.7\% & 5973 & \ppcell{53.4\%}{-5.3} & \relcell{1950}{-67.4} & \ppcell{60.0\%}{+1.3} & \relcell{3520}{-41.1} & \ppcell{56.9\%}{-1.8} & \relcell{5106}{-14.5} \\
Qwen3 14B & 61.5\% & 4597 & \ppcell{51.8\%}{-9.7} & \relcell{1500}{-67.4} & \ppcell{57.9\%}{-3.7} & \relcell{2111}{-54.1} & \ppcell{61.3\%}{-0.2} & \relcell{3665}{-20.3} \\
Qwen3.8 27B & 83.2\% & 4303 & \ppcell{79.2\%}{-4.0} & \relcell{2740}{-36.3} & \ppcell{84.0\%}{+0.8} & \relcell{4102}{-4.7} & \ppcell{85.3\%}{+2.1} & \relcell{4324}{+0.5} \\
R1 Qwen 7B & 49.1\% & 6686 & \ppcell{23.6\%}{-25.5} & \relcell{1305}{-80.5} & \ppcell{49.6\%}{+0.5} & \relcell{4037}{-39.6} & \ppcell{46.7\%}{-2.4} & \relcell{5953}{-11.0} \\
R1 Llama 8B & 47.7\% & 7424 & \ppcell{32.1\%}{-15.6} & \relcell{1771}{-76.2} & \ppcell{49.6\%}{+1.9} & \relcell{4526}{-39.0} & \ppcell{48.6\%}{+0.9} & \relcell{5816}{-21.7} \\
R1 Qwen3 8B & 60.8\% & 9084 & \ppcell{48.1\%}{-12.6} & \relcell{2398}{-73.6} & \ppcell{61.9\%}{+1.2} & \relcell{6595}{-27.4} & \ppcell{59.8\%}{-1.0} & \relcell{8877}{-2.3} \\
Claude Opus 5 & 89.5\% & 689 & \ppcell{88.7\%}{-0.9} & \relcell{564}{-18.2} & \ppcell{89.2\%}{-0.4} & \relcell{521}{-24.3} & \ppcell{87.6\%}{-2.0} & \relcell{686}{-0.5} \\
\midrule[0.08em]
\rowcolor{gray!12}
Average & 64.7\% & 5802 & $-10.5$ & \multicolumn{1}{l}{$-60.9\%$} & \multicolumn{1}{|l}{$+0.2$} & $-32.8\%$ & $-0.5$ & $-8.9\%$ \\
\midrule
\multicolumn{9}{l}{\textit{GSM8K}} \\
\midrule
Qwen3 4B T. & 95.2\% & 1151 & \ppcell{91.9\%}{-3.3} & \relcell{472}{-59.0} & \ppcell{93.4\%}{-1.8} & \relcell{467}{-59.5} & \ppcell{94.9\%}{-0.3} & \relcell{793}{-31.2} \\
Qwen3 8B & 95.4\% & 769 & \ppcell{95.2\%}{-0.2} & \relcell{371}{-51.8} & \ppcell{95.0\%}{-0.4} & \relcell{279}{-63.7} & \ppcell{95.1\%}{-0.3} & \relcell{537}{-30.1} \\
Qwen3 14B & 96.1\% & 587 & \ppcell{95.9\%}{-0.2} & \relcell{253}{-56.9} & \ppcell{95.9\%}{-0.2} & \relcell{200}{-66.0} & \ppcell{95.7\%}{-0.3} & \relcell{314}{-46.6} \\
Qwen3.8 27B & 97.4\% & 263 & \ppcell{97.5\%}{+0.1} & \relcell{200}{-24.0} & \ppcell{97.5\%}{+0.1} & \relcell{158}{-40.1} & \ppcell{97.5\%}{0.0} & \relcell{157}{-40.1} \\
R1 Qwen 7B & 91.4\% & 834 & \ppcell{87.5\%}{-3.9} & \relcell{187}{-77.6} & \ppcell{90.8\%}{-0.7} & \relcell{308}{-63.1} & \ppcell{91.6\%}{+0.1} & \relcell{824}{-1.2} \\
R1 Llama 8B & 89.5\% & 685 & \ppcell{81.8\%}{-7.8} & \relcell{278}{-59.4} & \ppcell{89.3\%}{-0.2} & \relcell{672}{-1.9} & \ppcell{89.4\%}{-0.1} & \relcell{797}{+16.4} \\
R1 Qwen3 8B & 92.8\% & 2377 & \ppcell{91.8\%}{-1.0} & \relcell{795}{-66.6} & \ppcell{93.0\%}{+0.2} & \relcell{911}{-61.7} & \ppcell{92.8\%}{0.0} & \relcell{2163}{-9.0} \\
\midrule[0.08em]
\rowcolor{gray!12}
Average & 94.0\% & 952 & $-2.3$ & \multicolumn{1}{l}{$-56.5\%$} & \multicolumn{1}{|l}{$-0.4$} & $-50.8\%$ & $-0.1$ & $-20.3\%$ \\
\midrule
\multicolumn{9}{l}{\textit{MATH-500}} \\
\midrule
Qwen3 4B T. & 97.6\% & 4064 & \ppcell{96.4\%}{-1.1} & \relcell{2148}{-47.2} & \ppcell{97.7\%}{+0.1} & \relcell{3716}{-8.6} & \ppcell{96.8\%}{-0.8} & \relcell{3605}{-11.3} \\
Qwen3 8B & 95.9\% & 3449 & \ppcell{95.6\%}{-0.3} & \relcell{1751}{-49.2} & \ppcell{95.1\%}{-0.8} & \relcell{2173}{-37.0} & \ppcell{95.9\%}{+0.1} & \relcell{2775}{-19.5} \\
Qwen3 14B & 96.9\% & 2604 & \ppcell{95.9\%}{-1.0} & \relcell{1375}{-47.2} & \ppcell{96.6\%}{-0.2} & \relcell{2454}{-5.8} & \ppcell{96.6\%}{-0.2} & \relcell{2444}{-6.2} \\
Qwen3.8 27B & 98.1\% & 1014 & \ppcell{96.7\%}{-1.5} & \relcell{748}{-26.2} & \ppcell{97.8\%}{-0.4} & \relcell{943}{-7.1} & \ppcell{97.2\%}{-0.9} & \relcell{994}{-2.0} \\
R1 Qwen 7B & 93.0\% & 2383 & \ppcell{69.8\%}{-23.2} & \relcell{1658}{-30.4} & \ppcell{93.2\%}{+0.2} & \relcell{2205}{-7.5} & \ppcell{93.4\%}{+0.4} & \relcell{2381}{-0.1} \\
R1 Llama 8B & 88.7\% & 3019 & \ppcell{79.4\%}{-9.3} & \relcell{1569}{-48.0} & \ppcell{88.5\%}{-0.2} & \relcell{2948}{-2.3} & \ppcell{88.8\%}{+0.2} & \relcell{3333}{+10.4} \\
R1 Qwen3 8B & 96.8\% & 6767 & \ppcell{93.0\%}{-3.8} & \relcell{3705}{-45.2} & \ppcell{96.6\%}{-0.2} & \relcell{6608}{-2.4} & \ppcell{96.7\%}{-0.2} & \relcell{6657}{-1.6} \\
\midrule[0.08em]
\rowcolor{gray!12}
Average & 95.3\% & 3329 & $-5.7$ & \multicolumn{1}{l}{$-41.9\%$} & \multicolumn{1}{|l}{$-0.2$} & $-10.1\%$ & $-0.2$ & $-4.3\%$ \\
\midrule
\multicolumn{9}{l}{\textit{Integrals}} \\
\midrule
Qwen3 4B T. & 99.9\% & 1914 & \ppcell{99.2\%}{-0.7} & \relcell{1029}{-46.3} & \ppcell{99.7\%}{-0.2} & \relcell{1411}{-26.3} & \ppcell{99.6\%}{-0.3} & \relcell{908}{-52.6} \\
Qwen3 8B & 99.7\% & 2887 & \ppcell{98.8\%}{-0.9} & \relcell{946}{-67.2} & \ppcell{99.8\%}{+0.1} & \relcell{2717}{-5.9} & \ppcell{98.8\%}{-0.9} & \relcell{1192}{-58.7} \\
Qwen3 14B & 99.8\% & 2321 & \ppcell{99.0\%}{-0.8} & \relcell{670}{-71.1} & \ppcell{99.5\%}{-0.3} & \relcell{1335}{-42.5} & \ppcell{99.6\%}{-0.2} & \relcell{756}{-67.4} \\
Qwen3.8 27B & 99.8\% & 724 & \ppcell{98.8\%}{-1.0} & \relcell{564}{-22.1} & \ppcell{99.5\%}{-0.3} & \relcell{624}{-13.8} & \ppcell{99.6\%}{-0.2} & \relcell{550}{-24.0} \\
R1 Qwen 7B & 96.4\% & 2519 & \ppcell{90.8\%}{-5.5} & \relcell{547}{-78.3} & \ppcell{96.0\%}{-0.4} & \relcell{2277}{-9.6} & \ppcell{96.3\%}{0.0} & \relcell{2291}{-9.0} \\
R1 Llama 8B & 93.7\% & 2462 & \ppcell{86.7\%}{-7.0} & \relcell{613}{-75.1} & \ppcell{93.1\%}{-0.6} & \relcell{2161}{-12.2} & \ppcell{93.6\%}{-0.1} & \relcell{1926}{-21.8} \\
R1 Qwen3 8B & 98.5\% & 4082 & \ppcell{96.7\%}{-1.8} & \relcell{1339}{-67.2} & \ppcell{97.9\%}{-0.6} & \relcell{4344}{+6.4} & \ppcell{98.5\%}{0.0} & \relcell{2701}{-33.8} \\
\midrule[0.08em]
\rowcolor{gray!12}
Average & 98.3\% & 2416 & $-2.5$ & \multicolumn{1}{l}{$-61.0\%$} & \multicolumn{1}{|l}{$-0.3$} & $-14.8\%$ & $-0.3$ & $-38.2\%$ \\
\midrule
\multicolumn{9}{l}{\textit{OneWordQA}} \\
\midrule
Qwen3 4B T. & 92.0\% & 668 & \ppcell{92.1\%}{+0.1} & \relcell{383}{-42.6} & \ppcell{91.7\%}{-0.3} & \relcell{179}{-73.2} & \ppcell{91.8\%}{-0.2} & \relcell{185}{-72.4} \\
Qwen3 8B & 91.7\% & 212 & \ppcell{92.0\%}{+0.3} & \relcell{71}{-66.7} & \ppcell{92.1\%}{+0.4} & \relcell{46}{-78.2} & \ppcell{92.2\%}{+0.5} & \relcell{46}{-78.1} \\
Qwen3 14B & 92.4\% & 207 & \ppcell{92.8\%}{+0.4} & \relcell{73}{-64.6} & \ppcell{93.2\%}{+0.7} & \relcell{48}{-76.6} & \ppcell{93.3\%}{+0.9} & \relcell{47}{-77.1} \\
Qwen3.8 27B & 93.3\% & 71 & \ppcell{93.4\%}{+0.1} & \relcell{40}{-43.8} & \ppcell{93.4\%}{+0.2} & \relcell{36}{-50.0} & \ppcell{93.4\%}{+0.2} & \relcell{35}{-51.3} \\
R1 Qwen 7B & 72.0\% & 454 & \ppcell{76.5\%}{+4.4} & \relcell{105}{-76.8} & \ppcell{76.5\%}{+4.4} & \relcell{98}{-78.3} & \ppcell{77.1\%}{+5.0} & \relcell{89}{-80.4} \\
R1 Llama 8B & 78.0\% & 484 & \ppcell{84.0\%}{+6.0} & \relcell{86}{-82.2} & \ppcell{83.5\%}{+5.5} & \relcell{74}{-84.6} & \ppcell{84.0\%}{+6.0} & \relcell{81}{-83.3} \\
\color{gray}R1 Qwen3 8B & \color{gray}81.5\% & \color{gray}328 & \color{gray}\ppcell{28.6\%}{-52.9} & \color{gray}\relcell{136}{-58.6} & \color{gray}\ppcell{82.1\%}{+0.6} & \color{gray}\relcell{354}{+7.8} & \color{gray}\ppcell{82.3\%}{+0.8} & \color{gray}\relcell{322}{-1.7} \\
\midrule[0.08em]
\rowcolor{gray!12}
Average & 86.6\% & 349 & $+1.9$ & \multicolumn{1}{l}{$-62.8\%$} & \multicolumn{1}{|l}{$+1.8$} & $-73.5\%$ & $+2.1$ & $-73.8\%$ \\
\midrule
\multicolumn{9}{l}{\textit{StrategyQA}} \\
\midrule
Qwen3 4B T. & 63.2\% & 836 & \ppcell{67.7\%}{+4.5} & \relcell{509}{-39.1} & \ppcell{68.2\%}{+5.0} & \relcell{248}{-70.4} & \ppcell{67.4\%}{+4.1} & \relcell{246}{-70.6} \\
Qwen3 8B & 70.1\% & 394 & \ppcell{72.1\%}{+1.9} & \relcell{96}{-75.6} & \ppcell{71.8\%}{+1.7} & \relcell{57}{-85.4} & \ppcell{70.8\%}{+0.7} & \relcell{52}{-86.8} \\
Qwen3 14B & 70.0\% & 366 & \ppcell{73.2\%}{+3.3} & \relcell{84}{-77.1} & \ppcell{73.5\%}{+3.5} & \relcell{60}{-83.6} & \ppcell{73.4\%}{+3.4} & \relcell{54}{-85.3} \\
Qwen3.8 27B & 79.4\% & 224 & \ppcell{80.2\%}{+0.8} & \relcell{151}{-32.4} & \ppcell{81.8\%}{+2.4} & \relcell{145}{-35.0} & \ppcell{81.0\%}{+1.6} & \relcell{144}{-35.5} \\
R1 Qwen 7B & 56.9\% & 651 & \ppcell{58.9\%}{+2.0} & \relcell{84}{-87.1} & \ppcell{58.5\%}{+1.6} & \relcell{57}{-91.2} & \ppcell{57.8\%}{+0.8} & \relcell{47}{-92.8} \\
R1 Llama 8B & 58.9\% & 446 & \ppcell{65.6\%}{+6.7} & \relcell{81}{-81.9} & \ppcell{65.2\%}{+6.3} & \relcell{95}{-78.7} & \ppcell{64.6\%}{+5.6} & \relcell{51}{-88.5} \\
R1 Qwen3 8B & 70.0\% & 606 & \ppcell{67.7\%}{-2.3} & \relcell{89}{-85.4} & \ppcell{71.1\%}{+1.1} & \relcell{425}{-29.9} & \ppcell{69.9\%}{-0.1} & \relcell{629}{+3.8} \\
\midrule[0.08em]
\rowcolor{gray!12}
Average & 66.9\% & 503 & $+2.4$ & \multicolumn{1}{l}{$-68.4\%$} & \multicolumn{1}{|l}{$+3.1$} & $-67.7\%$ & $+2.3$ & $-65.1\%$ \\
\midrule
\multicolumn{9}{l}{\textit{MMLU-Redux}} \\
\midrule
Qwen3 4B T. & 84.7\% & 1944 & \ppcell{82.9\%}{-1.8} & \relcell{527}{-72.9} & \ppcell{85.3\%}{+0.6} & \relcell{527}{-72.9} & \ppcell{84.9\%}{+0.2} & \relcell{1663}{-14.5} \\
Qwen3 8B & 87.2\% & 921 & \ppcell{84.0\%}{-3.3} & \relcell{306}{-66.7} & \ppcell{86.1\%}{-1.1} & \relcell{285}{-69.0} & \ppcell{86.9\%}{-0.3} & \relcell{922}{+0.2} \\
Qwen3 14B & 89.0\% & 718 & \ppcell{85.8\%}{-3.3} & \relcell{212}{-70.5} & \ppcell{88.2\%}{-0.9} & \relcell{319}{-55.6} & \ppcell{88.3\%}{-0.7} & \relcell{678}{-5.5} \\
Qwen3.8 27B & 92.4\% & 488 & \ppcell{91.9\%}{-0.5} & \relcell{262}{-46.4} & \ppcell{91.8\%}{-0.6} & \relcell{196}{-59.8} & \ppcell{92.1\%}{-0.3} & \relcell{352}{-27.9} \\
R1 Qwen 7B & 70.1\% & 1079 & \ppcell{60.6\%}{-9.5} & \relcell{397}{-63.2} & \ppcell{69.7\%}{-0.3} & \relcell{601}{-44.3} & \ppcell{70.4\%}{+0.3} & \relcell{1127}{+4.5} \\
R1 Llama 8B & 75.5\% & 1458 & \ppcell{58.6\%}{-16.9} & \relcell{431}{-70.4} & \ppcell{75.9\%}{+0.3} & \relcell{896}{-38.5} & \ppcell{75.3\%}{-0.2} & \relcell{1430}{-1.9} \\
R1 Qwen3 8B & 81.6\% & 1711 & \ppcell{64.5\%}{-17.1} & \relcell{279}{-83.7} & \ppcell{82.1\%}{+0.5} & \relcell{1273}{-25.6} & \ppcell{82.2\%}{+0.6} & \relcell{1714}{+0.2} \\
\midrule[0.08em]
\rowcolor{gray!12}
Average & 82.9\% & 1188 & $-7.5$ & \multicolumn{1}{l}{$-67.7\%$} & \multicolumn{1}{|l}{$-0.2$} & $-52.2\%$ & $-0.1$ & $-6.4\%$ \\
\bottomrule
\end{tabular}}
\end{table}
Table~\ref{tab:batching_results_full} extends Table~\ref{tab:batching_results} to GSM8K, Integrals, OneWordQA and StrategyQA. We leave out AIME, where nearly every task degrades beyond the 0.5\% budget already at $k=3$, so fragility-aware batching falls back to $k=1$ throughout. We note that the pattern observed in the selected datasets in the main text also holds for the other datasets we tested. Random batching at $k{=}6$ saves many tokens but loses accuracy, most visibly on reasoning-heavy datasets such as GPQA-Diamond, MMLU-Redux and MATH-500. Fragility-aware batching keeps accuracy close to the isolated baseline while still using substantially fewer tokens, and the linear predictor recovers this gain partially without access to measured fragility. Claude Opus~5 appears only on GPQA-Diamond, and to limit cost it is measured with fewer trials than the open-weight models: random batching rests on 5 trials per task instead of 10, the measured-fragility assignment on 5 isolated and 5 composed trials per task instead of 10 each, and the predictor's labels on 10 isolated trials and 5 per composition size.\looseness=-1

R1 Qwen3 8B is the one model that collapses on OneWordQA under random batching, from $81.5\%$ to $28.6\%$. Contrary to asking questions in isolation, in composition R1 Qwen3 8B often fails to adhere to the answering contract under composition, leading to parsing errors.\looseness=-1

When testing on the Integrals dataset, fragility-aware batching with held-out predictions from the linear predictor outperforms the measured variant. This is because even comparably fragile integrals lose very little accuracy, so estimates from a limited number of observations are noisy. The trained predictor also has access to the structure of each integral, which makes it more robust to this measurement noise.

On OneWordQA and StrategyQA, batching even raises accuracy for most models, by up to $6.7$ points (R1 Llama 8B on StrategyQA under random batching). We observe that both datasets need little reasoning. OneWordQA only asks the model to recall a fact, and StrategyQA to chain a few commonsense inferences. On such tasks, models seem to think more than the task needs, which can degrade performance too, consistent with \citet{srivastava2026constraint}.\looseness=-1

\providecommand{\relcell}[2]{{\small #1}\,{\scriptsize$(#2\%)$}}
\providecommand{\ppcell}[2]{{\small #1}\,{\scriptsize$(#2)$}}

\FloatBarrier
\subsection{Methodological details}

\subsubsection{Measuring per-task thinking length}
\label{app:trace-segmentation}
To measure how a model divides its thinking across the tasks of a composition, we prefix each task with a unique tag (\texttt{Q01}, \texttt{Q02}, \dots) and instruct the model to name the tag whenever it turns to that task. The tags mark where the reasoning on one task ends and the next begins. We assign every token after a tag to the corresponding task, up to the next tag or the end of the thinking trace. The thinking length $T_i(k)$ of task $t_i$ in a composition of size $k$ is the total number of tokens assigned to it, and the model may return to a task more than once. A task that is never named receives no tokens.\looseness=-1

\begin{quote}\small\ttfamily
<think> Q01: The dye emits light with an energy of 2.34 eV \dots\ maybe A.\\
Q02: A dual catalyst system for ethylene polymerization \dots\ tentatively D.\\
Back to Q01: 2.34 eV is about 530 nm, so the absorbed light \dots\ B.\\
Q03: \dots\ </think>
\end{quote}

\noindent Here the tokens after both \texttt{Q01} tags count towards $t_1$, while $t_2$ receives the tokens between its tag and the second \texttt{Q01}.\looseness=-1

This rule cannot reliably find where the trace for the last question in a composition ends, since no next tag marks its end, so its length is often overestimated. To fix this, we append an ungraded dummy question at the end of each composition.\looseness=-1

In our experiments, this tagging strategy works quite well: LLMs name every task in $99.5\%$ of all tagged thinking traces. GPQA-Diamond is the exception at $92.9\%$, mostly because R1 Qwen 7B names every task in only $61.5\%$ of its traces at $k \ge 3$.\looseness=-1

\subsubsection{Sampling parameters and context length}
\label{app:sampling-llms}

Every open-weight model is decoded with the same settings: temperature $0.6$ and top-$p$
$0.95$, with no top-$k$ cutoff and
no repetition penalty. These are the values Qwen recommends for its models in
thinking mode~\citep{yang2025qwen3technicalreport}, and the values DeepSeek recommends for the R1-distilled models. Each trial is an independent sample; we set no per-request seed. The context window is shared by the prompt and the model's response, including its thinking. It is $131{,}072$ tokens for Qwen3 8B, Qwen3 14B, Qwen3.8 27B and the three R1-distilled models, and $262{,}144$ for Qwen3 4B Thinking. Claude Opus~5 is queried through the OpenRouter API with reasoning effort set to medium, a $200{,}000$-token context window and at most $128{,}000$ output tokens, thinking included. With these limits, fewer than 1\% of responses reach the context limit in most datasets (Table~\ref{tab:context-parsing}).\looseness=-1

\begin{table}[t]
\centering
\small
\caption{Context limits and parsing failures in the composed sets. Each response has a token budget equal to the context window minus the prompt, capped at the model's maximum generation length; a response that exhausts it is truncated. Response tokens are the mean at the largest $k$. Truncation and parsing-failure rates are pooled over all models and $k$, with the worst model and $k$ in brackets. $^\dagger$Most unparseable responses come from R1 Qwen3 8B, which often breaks the answering contract under composition.}
\label{tab:context-parsing}

\begin{tabular}{lrrrrr}
\toprule
 & & \multicolumn{2}{c}{Response at largest $k$} & \multicolumn{2}{c}{Responses (\%)} \\
\cmidrule(lr){3-4}\cmidrule(lr){5-6}
Dataset & $k$ & tokens & of budget & hit limit & unparseable \\
\midrule
GSM8K & 20 & 2.8k & 3\% & 0.0 [0] & 0.2 [2] \\
MATH-500 & 20 & 19.6k & 29\% & 0.9 [10] & 2.4 [32] \\
AIME & 10 & 57.6k & 35\% & 4.8 [39] & 10.7 [47] \\
Integrals & 20 & 7.4k & 17\% & 0.0 [0] & 1.4 [15] \\
MMLU-Redux$^{\dagger}$ & 20 & 1.6k & 4\% & 0.2 [2] & 3.9 [44] \\
GPQA-Diamond & 10 & 8.9k & 8\% & 0.2 [1] & 2.3 [20] \\
OneWordQA & 20 & 0.8k & 2\% & 0.1 [0] & 0.1 [1] \\
StrategyQA & 20 & 1.1k & 1\% & 0.0 [0] & 0.0 [0] \\
\bottomrule
\end{tabular}%

\end{table}

\subsubsection{Correlating accuracy degradation across conditions}
\label{app:degradation-correlation}

\paragraph{Conditions and degradation variables.}
We define a \emph{condition} as a tuple $c = (m, k)$ of a model $m$ and a composition size $k$, e.g.\ $(\text{Qwen3 8B}, k=20)$. For a condition $A = (m_A, k_A)$ we write $\Delta_A$ for the random variable that assigns to each task $t_i$ its accuracy degradation $\Delta^{m_A}_i(k_A)$ under that condition, and analogously $\Delta_B$ for the second condition $B = (m_B, k_B)$ that we correlate against. All correlations below are computed over all tasks of a dataset $t_i \in \mathcal{D}$.\looseness=-1

\paragraph{Correlation and disattenuation.}
For each pair of conditions we compute the Pearson correlation $r$ between $\Delta_A$ and $\Delta_B$ over all tasks of a dataset. Because $\Delta_A$ and $\Delta_B$ are estimated from a finite number of trials, measurement noise attenuates $r$ toward zero. We therefore apply Spearman's correction for attenuation to every pair of conditions separately,
\begin{equation}
r^{*} = \frac{r}{\sqrt{r_{AA}\, r_{BB}}},
\end{equation}
where the reliability $r_{AA}$ is the test-retest correlation of $\Delta_A$, that is, the correlation between the accuracy drops at condition $A$ measured in the first and in the second pass. We then average $r^{*}$ arithmetically over all pairs of conditions and datasets that enter a reported value. We do not clip $r^{*}$.\looseness=-1

\paragraph{Avoiding shared measurement noise.}
When $A$ and $B$ share a model, both drops subtract the same isolated accuracy, so its estimation error would inflate their correlation. We therefore measure every task in two independent passes and take $\Delta_A$ from one pass and $\Delta_B$ from the other.\looseness=-1

\paragraph{Confidence intervals.}
We report half the width of a $95\%$ percentile interval from a bootstrap over tasks. In each of $5000$ replicates we resample the tasks of a dataset with replacement at the original size and recompute every correlation and every reliability on the resampled tasks.\looseness=-1

\subsubsection{Implementation details of the fragility-aware batching algorithm}
\label{app:fragility-aware-batching}

\paragraph{Estimating $\mathbb{E}[\Delta_i(k)]$.}
For the fragility-aware batching intervention in Section~\ref{sec:oracle-fix}, each task has 10 isolated trials and 10 trials at every composition size. We split these trials at random into two halves, A and B. For each $k$, an isotonic map from the drop measured on half A to the drop measured on half B gives $\mathbb{E}[\Delta_i(k)]$.\looseness=-1

\paragraph{Estimating $\mathrm{cost}_k$.}
The ratio $\mathrm{cost}_k$ does not depend on the individual task and is
estimated once per model and dataset as the mean number of output tokens
attributable to a single task inside a composition of size $k$, divided by the
mean number of output tokens used for isolated solving:
\begin{equation}
\mathrm{cost}_k = \frac{\frac{1}{k}\,\overline{T}(k)}{\overline{T}(1)},
\end{equation}
where $\overline{T}(k)$ is the mean total output length of a composition of size
$k$. In each experiment, we estimate it over the entire dataset.\looseness=-1

\paragraph{Selecting $\lambda^{*}$ by repeated cross-fitting.}
Per-task accuracy estimates are noisy and our pool of tasks is limited, so a $\lambda^{*}$ selected on the same data used to estimate $\mathbb{E}[\Delta_i(k)]$ would be optimistically biased. We therefore split the task pool into 5 outer folds and determine $\lambda^{*}$ for each fold only on its training part.\looseness=-1

Within a training slice, we split the tasks into 5 blocks, four times under different random shuffles, which gives 20 held-out blocks. For each block, we fit the isotonic maps on the remaining tasks, apply them to the block and assign every task in it a composition size by Equation~\ref{eq:lagrangian}. We do this for a grid of $121$ exchange rates ($\lambda = 0$ together with $120$ values spaced logarithmically over $[10^{-4}, 3.0]$) and record the accuracy and token ratio realized on half B.\looseness=-1

Averaging the 20 blocks results in one accuracy curve over $\lambda$, which we make non-increasing by isotonic regression, since a larger exchange rate can only move tasks to larger composition sizes. The term $\mathrm{SE}(\bar{\Delta}(\lambda))$ in Equation~\ref{eq:lambda} is the standard error of the realized accuracy across the 20 blocks. We select the cheapest rate whose accuracy, minus one standard error, stays above $\overline{\mathrm{acc}}(1) - B$, where $\overline{\mathrm{acc}}(1)$ is the mean isolated accuracy. If no rate clears this threshold, we take the rate with the highest accuracy minus one standard error.\looseness=-1

Finally, we fit the isotonic maps on the whole training slice, apply them to the accuracy drops measured on half A of the held-out fold and assign each of its tasks a composition size at $\lambda^{*}$. Only these held-out assignments enter the reported results.\looseness=-1

\paragraph{Forming compositions.}
The assignment yields a per-task upper bound $k_i(\lambda^{*})$ on the size of the composition that task may be placed in, and compositions must be formed such that $|C| \le \min_{t_i \in C} k_i(\lambda^{*})$ for every composition $C$. Placing a task in a composition smaller than its assigned size is admissible, since its expected drop is then no larger, so the realized mean drop should not exceed the planned one. We use the greedy procedure in Algorithm~\ref{alg:batching}: proceeding from the smallest assigned size upwards, we form full compositions from the tasks assigned to the current size and fill any remaining partial composition with tasks assigned to strictly larger sizes, which are by construction more robust and whose bound is therefore not violated. Filler tasks come from the nearest larger size first and are drawn at random within it. Compositions are formed separately within each outer test fold, so tasks from different folds never share a prompt.\looseness=-1

\begin{algorithm}[t]
\caption{Forming compositions from a size assignment}
\label{alg:batching}
\small
\begin{algorithmic}[1]
\Require Tasks $t_1, \dots, t_n$ with assigned sizes $k_i \in \mathcal{K}$
\Ensure A set of compositions $\mathcal{B}$
\State $\mathcal{B} \gets \emptyset$; \quad $U \gets \{t_1, \dots, t_n\}$
\For{$k \in \mathcal{K}$ in increasing order}
  \State $P \gets \{\, t_i \in U : k_i = k \,\}$
  \While{$|P| \ge k$}
    \State remove $k$ randomly chosen tasks from $P$ and from $U$, and add them as composition $\mathcal{B} \leftarrow \mathcal{B} \cup\{(t_1, ..., t_k)\}$
  \EndWhile
  \If{$P \neq \emptyset$}
    \State $F \gets$ up to $k - |P|$ tasks from $\{\, t_i \in U : k_i > k \,\}$,
      nearest larger size first, at random within a size
    \State remove $P \cup F$ from $U$ and add $P \cup F$ as a composition to $\mathcal{B}$
  \EndIf
\EndFor
\State \Return $\mathcal{B}$
\end{algorithmic}
\end{algorithm}

If fewer than $k - |P|$ tasks with a larger assigned size remain, the leftover tasks form a composition smaller than $k$. Since filler tasks are consumed from the pools of larger assigned sizes, the realized mean composition size is slightly below the assigned one, so the reported token savings are a conservative estimate of what the assignment permits.\looseness=-1

\subsubsection{Predicting fragility from the task}
\label{app:probe}

\paragraph{Architecture.}
The predictor takes the task description and embeds it with Qwen3 Embedding 8B
using last-token pooling. A linear head maps the embedding to a single logit
$\ell_i(k_1)$, the logit for the smallest composition size $k_1 = 3$. The logits for the
remaining sizes are obtained by accumulating non-negative offsets along increasing k,\looseness=-1

\begin{equation}
\ell_i(k_1) = w^{\top} e_i + b,
\qquad
\ell_i(k_{j}) = \ell_i(k_{j-1}) - \mathrm{softplus}(\theta_j),
\quad j = 2, \dots, J,
\end{equation}

where $e_i$ is the embedding of task $t_i$, $k_1 < \dots < k_J$ are
the composition sizes in $\mathcal{K} \setminus \{1\} = \{3, 6, 10, 20\}$ ($\{3, 6, 10\}$ for GPQA-Diamond), and $\theta_j$ are
trainable scalars shared across all tasks. The softplus constrains every offset to
be non-negative, so the predicted probability is non-increasing in $k$ by
construction, matching the behavior established in
Section~\ref{sec:results}. Accuracy measurements under composition are expensive, so the predictor has to learn from little data. We therefore use a linear head, which has few parameters and builds monotonicity in $k$ into its architecture.\looseness=-1

\paragraph{Training.}
For every task and every $k \in \mathcal{K} \setminus \{1\}$ we form a binary target indicating
that the task performs no worse under composition than in isolation, using the
$10$ composed and $10$ isolated trials measured for that task. The head is trained
with a class-balanced binary cross-entropy loss averaged over all tasks and composition sizes, so a single
coefficient vector is shared across the ladder and the sizes differ only through
the learned offsets. We optimize with AdamW at a learning rate of $3 \times
10^{-3}$, weight decay $1.0$ and a cosine schedule for $2000$ steps with batch size $64$ and gradient
clipping at $1.0$. Every $100$ steps we evaluate the head on its inner held-out fold and keep the best checkpoint.\looseness=-1

\paragraph{Nested cross-fitting.}
The predictor and the exchange rate $\lambda^{*}$ must both be estimated without
access to the data on which the intervention is evaluated, so we nest the fitting
of the head inside the cross-fitting scheme of Appendix~\ref{app:fragility-aware-batching}.\looseness=-1

We split the dataset into $5$ outer folds. For each outer fold, the $4/5$ training
slice is split again into $4$ inner folds; one head is trained on each inner training
split and predicts on the corresponding inner test split. Concatenating these
predictions yields an out-of-training prediction for every task in the outer
training slice, which takes the place of the half-A accuracy drops used in
Appendix~\ref{app:fragility-aware-batching}. The isotonic maps then run from this prediction to the accuracy drop measured on all $10$ trials. On the held-out fold, we average the predicted accuracy drops of the four heads, and the remainder of the intervention is unchanged.\looseness=-1

\lstdefinestyle{prompt}{
  basicstyle=\ttfamily\footnotesize,
  breaklines=true,
  breakatwhitespace=true,
  columns=fullflexible,
  keepspaces=true,
  showstringspaces=false,
  upquote=false,
  escapeinside={(*@}{@*)},
}
\newtcblisting{promptbox}[1]{
  listing only,
  breakable,
  colback=gray!4,
  colframe=gray!55,
  boxrule=0.4pt,
  arc=1pt,
  left=4pt, right=4pt, top=2pt, bottom=2pt,
  title={\small #1},
  coltitle=black,
  colbacktitle=gray!18,
  listing options={style=prompt},
}
\newcommand{\ph}[1]{$\langle$\textit{#1}$\rangle$}

\subsubsection{Further details on datasets}
\label{app:datasets}

\paragraph{Indefinite integrals.}
We generate the integral pool ourselves, to ensure that every problem is known to be solvable and simple enough for most small open-weight models to solve confidently.
It holds 1{,}000 atomic integrands, 200 from each of five function classes:
\begin{itemize}
  \item \emph{polynomial}: $a\,x^{n}$;
  \item \emph{exponential}: $a\,e^{bx}$;
  \item \emph{trigonometric}: $a\,f(bx+c)$ with $f\in\{\sin,\cos,\tan\}$;
  \item \emph{logarithmic and rational}: $a\ln(bx+c)$ or $a/(bx+c)$;
  \item \emph{integration by parts}: $a\,x\,f(bx+c)$ with
        $f\in\{\exp,\sin,\cos,\ln,\arctan\}$.
\end{itemize}
The coefficient $a$ is a nonzero integer in $[-5,5]$. The exponent $n$ and the
frequency $b$ are nonzero integers in $[-15,15]$, and the shift $c$ is an integer in
$[-15,15]$. All parameters are drawn with a fixed seed. Structural duplicates are
discarded.\looseness=-1

An integrand is kept only if SymPy finds an elementary antiderivative for it, and only
if that antiderivative passes the same checker that grades the models. The checker works
symbolically rather than by string matching. It parses the boxed LaTeX answer,
differentiates it, and tests whether the derivative equals the integrand, first
structurally, then numerically at random points, and finally with a symbolic proof
under a timeout. An additive constant is accepted, but any other free symbol is
rejected. Answers that cannot be parsed, or whose check times out, are recorded
separately rather than counted as wrong.\looseness=-1

\paragraph{OneWordQA.}
OneWordQA is a set of 1{,}548 questions that we generated using Claude Opus 5 and checked manually. They cover 23
categories of simple factual lookups whose answers leave little room for
interpretation. Examples include country capitals, dialing and ISO codes, chemical
symbols and atomic numbers, Roman numerals, polygon sides and the number of players
per team in a sport. Each answer is a single word, a number or a short code.
Grading normalizes only how an answer is written (case, diacritics and surrounding
punctuation). Where a question has more than one
correct answer, the alternatives are listed explicitly in the dataset.\looseness=-1

\paragraph{Benchmarks.}
The six public benchmarks are loaded from the Hugging Face Hub at pinned revisions. We use the error-free questions of MMLU-Redux, drawing the same number from each of 19 subjects (74 per subject, 1{,}406 questions), and keep the answer options in their published order. GPQA-Diamond contributes 196 of its 198 questions; we drop the two that repeat an answer option. From GSM8K we take the 1{,}319 problems of the test split, from MATH-500 all 500 problems, and from AIME the 60 problems of 2024 and 2025. StrategyQA contributes the 687 questions of its test split, and an answer counts as correct when it is the gold yes or no.\looseness=-1

\paragraph{Prompts.}
\label{app:prompts}
We now list the prompt every model receives, for each of the eight datasets used in
the main experiments: indefinite integrals, MMLU-Redux, GPQA-Diamond, OneWordQA,
GSM8K, MATH-500, AIME and StrategyQA.\looseness=-1

Conventions shared by all datasets:
\begin{itemize}
  \item An atomic task is
        treated as a composition of size $k=1$ and uses the same template
        as a $k$-task composition, with the plural wording made singular.
  \item The prompt is sent as a single
        user message through the model's chat template, with no few-shot
        examples and no system message.
  \item Models must answer in the order the tasks are shown, which is
        the order in which we grade them.

\end{itemize}

\textbf{Indefinite integrals}
\label{app:prompts:integrals}

A composition of $k$ atomic integrands is shown as one integral whose integrand
is their sum, rendered in LaTeX by SymPy. \ph{integral} is that LaTeX string,
for example:\looseness=-1
\begin{quote}
\verb|\int 3 x^{2} + 2 \sin{\left(4 x + 1 \right)} \, dx|.
\end{quote}

The prompt shows the whole sum unchanged, but the model reports one
antiderivative per term. The terms are listed in the order in which they appear in the rendered integral. Answer $i$ is checked symbolically against term $i$.\looseness=-1

\begin{promptbox}{Indefinite integrals}
Find an antiderivative for each term of the following indefinite integral: (*@\ph{integral}@*).

The integrand is a sum of (*@\ph{k}@*) terms, listed below in the order they appear in the integral:
1. \int (*@\ph{term 1}@*) \, dx
2. \int (*@\ph{term 2}@*) \, dx
...
(*@\ph{k}@*). \int (*@\ph{term k}@*) \, dx
Do not add the antiderivatives together, and do not give an antiderivative of the whole sum. Solve each term on its own and report (*@\ph{k}@*) separate antiderivatives, one per term, in the order above.

Give symbolic answers only in standard LaTeX format. Do not give a decimal approximation. The constant of integration may be omitted.
Return your answers in the following format in the last (*@\ph{k}@*) lines, with no additional text: one line per term, in the order above, each line holding exactly one \boxed{}.

1. \boxed{<antiderivative of term 1>}
2. \boxed{<antiderivative of term 2>}
...
(*@\ph{k}@*). \boxed{<antiderivative of term (*@\ph{k}@*)>}

Before you finish, re-read your answer block and check that it holds exactly (*@\ph{k}@*) lines, each with exactly one \boxed{}, in the order the terms are listed above.
\end{promptbox}

\textbf{MMLU-Redux and GPQA (multiple choice)}
\label{app:prompts:mcq}

Both datasets use the same multiple-choice layout. Every question is followed by
its lettered options, and consecutive questions are separated by a blank line:\looseness=-1

\begin{promptbox}{Question block (\ph{question block}) for both datasets}
1. (*@\ph{question 1}@*)
A. (*@\ph{option A}@*)
B. (*@\ph{option B}@*)
C. (*@\ph{option C}@*)
D. (*@\ph{option D}@*)

2. (*@\ph{question 2}@*)
A. ...
\end{promptbox}

In GPQA, the four options of each question are shuffled once, with a fixed seed,
when the dataset is built, so the correct answer does not always sit in the same
position. \ph{example} is the letters offered, cycled to length $k$ (for
example \texttt{A, B, C, D, A, B} for $k=6$). For $k=1$ it is \texttt{A}.\looseness=-1

\begin{promptbox}{MMLU-Redux ($k>1$)}
Answer the following multiple-choice questions.

(*@\ph{question block}@*)

For each question, give only the letter of the single best option.
End your response with a line of the following form, and write nothing after it: a comma-separated list of (*@\ph{k}@*) letters, one per question, in the order the questions were asked.
Answer: (*@\ph{example}@*)
\end{promptbox}

The GPQA prompt is the MMLU-Redux prompt word for word, with one extra closing
sentence. GPQA questions are long and hard, and its thinking traces are several times longer than those on MMLU-Redux. After such long reasoning, models often lose the answer format they were given at the start, so we ask them to re-check their answer line. The sentence concerns only the format of the answer line, so it does not change what the two prompts ask about the questions:\looseness=-1

\begin{promptbox}{GPQA: closing sentence appended to the MMLU-Redux prompt}
Before you finish, re-read that final line and check that it holds exactly (*@\ph{k}@*) letters, in the order the questions were asked, with nothing written after them.
\end{promptbox}

For $k=1$ this sentence reads ``\dots holds exactly 1 letter, with nothing
written after it.''\looseness=-1

\textbf{OneWordQA and StrategyQA}
\label{app:prompts:onewordqa}

Both datasets use the same layout. Each question sits on a single line, numbered,
and consecutive questions are separated by a single newline (\ph{question block}).
The answer is a numbered block of $k$ lines under the heading \texttt{Answer:}.\looseness=-1

\begin{promptbox}{OneWordQA ($k>1$)}
Answer the following questions. Each answer is short -- usually a single word or a number.

(*@\ph{question block}@*)

Give the answer only: no explanation, and no units unless the question asks for them.
End your response with an answer block: the line "Answer:" followed by exactly (*@\ph{k}@*) lines, one answer per line, numbered, in the order the questions were asked. Write nothing after the block.

Answer:
1. <your answer to question 1>
2. <your answer to question 2>
...
(*@\ph{k}@*). <your answer to question (*@\ph{k}@*)>
Before you finish, re-read your answer block and check that it holds exactly (*@\ph{k}@*) answer lines, in the order the questions were asked, with nothing written after them.
\end{promptbox}

The StrategyQA prompt is the OneWordQA prompt word for word, apart from two
sentences that restrict the answer to yes or no:\looseness=-1

\begin{promptbox}{StrategyQA: the two sentences that differ from OneWordQA}
Answer the following questions. Each answer is either yes or no.
...
Give the answer only, as the single word yes or no: no explanation, and no answer other than those two words.
\end{promptbox}

The instruction is not enforced strictly. Case and punctuation are ignored, and
\texttt{true} and \texttt{false} are accepted alongside \texttt{yes} and
\texttt{no}. For $k=1$ the first sentence reads ``Answer the following
question.''\looseness=-1

\textbf{GSM8K, MATH-500 and AIME (short mathematical answers)}
\label{app:prompts:shortmath}

The three datasets use the same answer contract as OneWordQA: a numbered block
of $k$ lines under the heading \texttt{Answer:}. The difference is in the
problems themselves, which are much longer (several sentences, displayed
equations, and for 42 MATH-500 problems an Asymptote figure). Consecutive
problems are therefore separated by a blank line rather than a single newline.
Blank lines inside a problem are collapsed when the pool is built, so a blank
line in \ph{problem block} always marks the start of the next problem. For
$k=1$ the answer block has the single line \texttt{1. <your answer>} (for
MATH-500, \verb|1. \boxed{<your answer>}|), and the closing sentence reads
``\dots holds exactly one answer line, with nothing written after it.'' The
three prompts differ only in their first line and in the sentence that describes
the answer format.\looseness=-1

\begin{promptbox}{GSM8K ($k>1$)}
Solve the following problems. Each answer is a single number.

(*@\ph{problem block}@*)

Give the final answer to each problem as a number only: no units, no currency symbol, no thousands separator and no explanation.
End your response with an answer block: the line "Answer:" followed by exactly (*@\ph{k}@*) lines, one answer per line, numbered, in the order the problems were asked. Write nothing after the block.

Answer:
1. <your answer to problem 1>
2. <your answer to problem 2>
...
(*@\ph{k}@*). <your answer to problem (*@\ph{k}@*)>
Before you finish, re-read your answer block and check that it holds exactly (*@\ph{k}@*) answer lines, in the order the problems were asked, with nothing written after them.
\end{promptbox}

In MATH-500 each answer is boxed. Many answers contain spaces or commas, such as
tuples, intervals, or \verb|10,\!080|, so the box marks exactly where an answer
begins and ends. The instruction ``in the form the problem asks for'' matters
because many MATH-500 problems set their own answer format (for example
``Express your answer as a common fraction''), and the reference answer follows
that format.\looseness=-1

\begin{promptbox}{MATH-500 ($k>1$)}
Solve the following problems.

(*@\ph{problem block}@*)

Give the final answer to each problem only, with no explanation. Write each answer in LaTeX, in the form the problem asks for, wrapped in \boxed{...}.
End your response with an answer block: the line "Answer:" followed by exactly (*@\ph{k}@*) lines, one answer per line, numbered, in the order the problems were asked. Write nothing after the block.

Answer:
1. \boxed{<your answer to problem 1>}
2. \boxed{<your answer to problem 2>}
...
(*@\ph{k}@*). \boxed{<your answer to problem (*@\ph{k}@*)>}
Before you finish, re-read your answer block and check that it holds exactly (*@\ph{k}@*) answer lines, in the order the problems were asked, with nothing written after them.
\end{promptbox}

\begin{promptbox}{AIME ($k>1$)}
Solve the following competition problems. Each answer is an integer between 0 and 999.

(*@\ph{problem block}@*)

Give the final answer to each problem as an integer between 0 and 999: digits only, no units, no thousands separator and no explanation.
End your response with an answer block: the line "Answer:" followed by exactly (*@\ph{k}@*) lines, one answer per line, numbered, in the order the problems were asked. Write nothing after the block.

Answer:
1. <your answer to problem 1>
2. <your answer to problem 2>
...
(*@\ph{k}@*). <your answer to problem (*@\ph{k}@*)>
Before you finish, re-read your answer block and check that it holds exactly (*@\ph{k}@*) answer lines, in the order the problems were asked, with nothing written after them.
\end{promptbox}

For $k\le 3$ the example answer block in all three prompts lists every line,
with no ellipsis.\looseness=-1

\end{document}